\documentclass{article}

\PassOptionsToPackage{numbers, compress}{natbib}

\usepackage{main}

\usepackage[utf8]{inputenc} %
\usepackage[T1]{fontenc}    %
\usepackage{hyperref}       %
\usepackage{url}            %
\usepackage{booktabs}       %
\usepackage{amsfonts}       %
\usepackage{nicefrac}       %
\usepackage{microtype}      %
\usepackage{xcolor, colortbl}         %
\usepackage{multirow}
\usepackage{tabularx}
\usepackage{subcaption}
\usepackage{graphicx}
\usepackage{enumitem}
\usepackage{algpseudocode}
\usepackage{array}
\usepackage{fontawesome5}

\definecolor{darkblue}{rgb}{0,0 ,0.6}
\hypersetup{
    colorlinks = true,
    citecolor = darkblue,
    linkcolor = {black}
}

\usepackage{mathtools} 
\usepackage{microtype} 
\usepackage{nicefrac}       
\usepackage{pifont}   
\usepackage{siunitx}
\usepackage{threeparttable}  

\definecolor{snowflakeColor}{RGB}{66,133,244} %
\definecolor{fireColor}{RGB}{255,87,34}        %

\usepackage[labelfont=bf, font=small, margin=0pt]{caption}

\title{Robust Multimodal Learning with Missing Modalities via Parameter-Efficient Adaptation}

\author{Md Kaykobad Reza$^1$, Ashley Prater-Bennette$^2$, M. Salman Asif$^1$ \\
$^1$ University of California Riverside, CA 92508, USA \\
$^2$ Air Force Research Laboratory, Rome, NY 13441, USA \\
\texttt{mreza025@ucr.edu, ashley.prater-bennette@us.af.mil, sasif@ucr.edu}
}

\begin{document}

\maketitle

\begin{abstract}
Multimodal learning seeks to utilize data from multiple sources to improve the overall performance of downstream tasks. It is desirable for redundancies in the data to make multimodal systems robust to missing or corrupted observations in some correlated modalities. However, we observe that the performance of several existing multimodal networks significantly deteriorates if one or multiple modalities are absent at test time. To enable robustness to missing modalities, we propose a simple and parameter-efficient adaptation procedure for pretrained multimodal networks. In particular, we exploit modulation of intermediate features to compensate for the missing modalities. We demonstrate that such adaptation can partially bridge performance drop due to missing modalities and outperform independent, dedicated networks trained for the available modality combinations in some cases. The proposed adaptation requires extremely small number of parameters (e.g., fewer than 1\% of the total parameters) and applicable to a wide range of modality combinations and tasks. We conduct a series of experiments to highlight the missing modality robustness of our proposed method on five different multimodal tasks across seven datasets. Our proposed method demonstrates versatility across various tasks and datasets, and outperforms existing methods for robust multimodal learning with missing modalities.
\end{abstract}

\section{Introduction}
Multimodal learning (MML) \cite{baltruvsaitis2018mml_survey1, xu2023mml_survey2} is a general framework for processing, combining, and understanding information from multiple, diverse data sources. Fusing knowledge from multiple modalities (e.g., text, images, audio, and sensor data) is expected to provide more accurate and reliable systems. In recent years, MML has achieved remarkable success in a wide range of applications, including image segmentation \cite{chen2020SAGate, wang2022tf, reza2024mmsformer}, captioning \cite{zhao2019mmcaption1, yu2019mmcaption2}, classification \cite{guillaumin2010mmclass1, roy2023mmclass2}, sentiment analysis \cite{soleymani2017mmsentiment1, kaur2022mmsentiment2}, and autonomous driving \cite{xiao2020mmdriv1, rizzoli2022mmdrive2}. 
In all these applications, one often encounters situations where some modalities are corrupted or missing due to hardware limitations/failures, privacy concerns or data acquisition cost/constraints. 
The ability to handle corrupt or missing modalities is thus crucial for the robustness and reliability of multimodal systems. %
However, most of the existing multimodal models are not designed to handle corrupt or missing modalities. The primary focus of this paper is to study and enhance robustness of existing multimodal models in different missing modality scenarios.

Recent studies \cite{ma2022robust_mmt1,hazarika2022robust_mmt2,cheng2023robust_mmt3} have shown that MML is not inherently robust to missing modalities and performance can drop significantly when modalities are missing at test time. Existing approaches for robust MML usually work for specific combinations of modalities they are trained for and tend to perform poorly when applied to untrained combinations. 
For instance, one approach is to adopt robust training strategies such as modality dropout during training \cite{neverova2015moddrop, hussen2020mdrop}, partial or full modality masking \cite{bachmann2022multimae, shin2023crm}, and knowledge distillation \cite{tarvainen2017meanteacher, maheshwari2023m3l}. These approaches either require specialized training strategies or utilize extra models/sub-networks to guide the underlying model. 
Another approach replaces uninformative tokens with aggregated informative tokens from different modalities or learns to predict tokens for the specific missing modalities \cite{wang2022tf,woo2023actionmae,shin2023crm}. Training such separate (independent) networks for every possible modality combination is not feasible specially when the number of input modalities is large. 
One recent approach for robust MML is to impute missing modalities from the available modalities \cite{yu2018gan2,sharma2019gan1,dorent2019vae}. Performance of these methods depend on the generation model that imputes the missing modalities. 

\begin{figure}[t]
  \centering
  \includegraphics[width=0.98\linewidth]{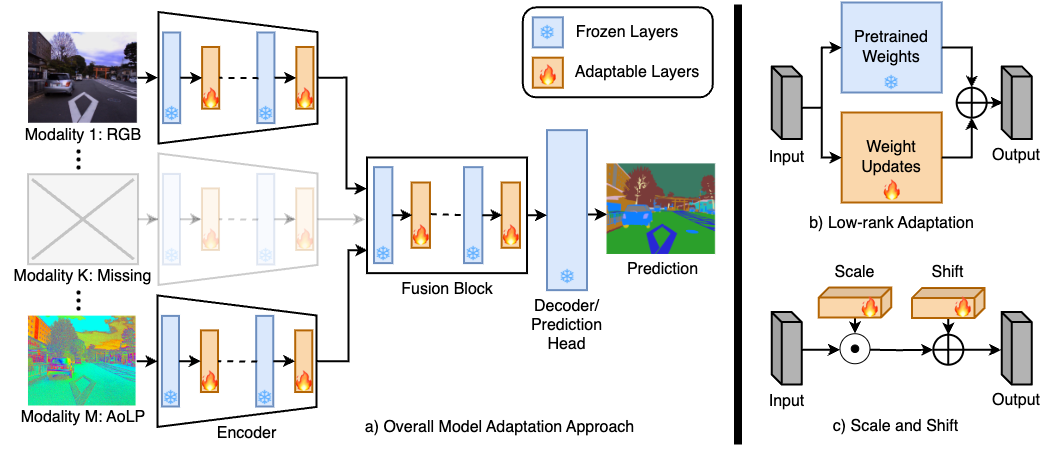} 
  \caption{a) Overview of our model adaptation approach for robust MML. A model pretrained on all the modalities is adapted using a small number of learnable parameters to handle different modality combinations. We insert adaptable layers after each layer of the encoders and the fusion block to learn the modulation as a function of the available input modalities to compensate for the missing modalities. The grayed-out branch (missing modality) is inactive and does not contribute to the output. b) Low-rank model adaption computes features using frozen weights and low-rank weight updates and combine them. c) Scale and shift feature adaptation transforms input by element-wise multiplication and addition.}
  \label{fig:intro-figure}
\end{figure}

In this paper, we propose a parameter-efficient approach to adapt existing multimodal networks to perform well on different missing modality scenarios. \textbf{Our main objective is to modify the network in a controllable manner as a function of available modalities. }
For instance, if a modality is missing, we seek to modify how the features from available modalities are extracted and fused for the inference. Instead of learning an independent network for each modality combination, our goal is to perform parameter-efficient adaptation. 
Figure~\ref{fig:intro-figure} illustrates our proposed method, where a given multimodal network can be adapted to arbitrary modality combinations by transforming the intermediate features of the available input modalities at different layers. 
To achieve parameter-efficient adaptation, we propose to use simple linear transformations such as scaling and shifting to modulate the intermediate features or low-rank increments of features to compensate for the missing modalities. Our method does not require retraining the entire model or any specialized training strategy. The adapted networks provide significant performance improvement over the multimodal networks trained with all modalities and tested with missing modalities. Performance of the adapted models is also comparable or better than the models that are exclusively trained for each input modality combination as shown in Table~\ref{tab:missing-modality-effect-on-cmnext}. We present a series of experiments to evaluate our method and compare with existing methods for robust MMLon five multimodal tasks across seven datasets (Section~\ref{sec:datasets-and-tasks}). We tested different parameter-efficient adaptation strategies and found intermediate feature modulation with scaling and shifting provides overall best performance, which is discussed in Section~\ref{sec:other-peft-comparison}. Our method shows significant performance improvement with less than 1\% additional learnable parameters as discussed in Section~\ref{sec:performance-vs-params}.

\textbf{Contributions.} The main contributions can be summarized as follows.
\begin{itemize}[leftmargin=*,noitemsep,topsep=0pt]
    \item We propose parameter-efficient adaptation procedure for multimodal learning that is robust to missing modalities. The adapted model can easily switch to different network states based on the available modalities with minimal latency, computational, or memory overhead. 
    \item The adapted networks provide notably improved performance with missing modalities when compared to models trained with all modalities and is comparable to or better than the networks trained for specific modality combinations (Table~\ref{tab:missing-modality-effect-on-cmnext}).
    \item Our approach is versatile and adaptable to a wide range of multimodal tasks, datasets and models. Detailed evaluations on different datasets and tasks show that our approach outperforms existing baseline methods and robust models designed for specific tasks and datasets (Section~\ref{sec:result-segmentation} - Section~\ref{sec:result-classification}). 
\end{itemize}

\section{Related Work}
\label{sec:related-works}
\hspace{\parindent}\textbf{Multimodal learning with missing modalities} has been studied for different applications in recent years. For instance, robustness in  vision-language tasks with multimodal transformers in \cite{ma2022robust_mmt1}, multimodal sentiment analysis in \cite{hazarika2022robust_mmt2},  multimodal classification  in \cite{lee2023map}, and  multimodal action recognition in \cite{woo2023actionmae}. These studies have shown that the task performance can drop significantly when modalities are missing during test time. 

\textbf{Robust training strategies} have been proposed to make models robust to different missing modalities. Such approaches include modality dropout during training  \cite{neverova2015moddrop, hussen2020mdrop}, unified representation learning \cite{lau2019urn}, and supervised contrastive learning \cite{gomaa2022mmcl}. Modality masking during training has become a popular choice for enhancing robustness. \cite{shin2023crm} utilized complementary random masking, \cite{he2022mae} used masked auto encoder, and \cite{ma2022robust_mmt1} applied masked cross-modal attention for enhancing robustness of the underlying model. \cite{hazarika2022robust_mmt2} proposed noisy perturbation of modalities during training for robust multimodal sentiment analysis. Recently,  \cite{li2023reviewer4} proposed uni-modal ensemble with modality drop and substitution augmentation during training to adapt to different missing modality scenarios.

\textbf{Design of robust models and fusion strategies} is another approach for robust MML. \cite{fan2023spidermesh} proposed a recursive meshing technique called SpiderMesh and \cite{shin2023crm} designed complementary random masking (CRM) and knowledge distillation based framework for robust RGB-thermal semantic segmentation. 
\cite{wang2022tf} proposed TokenFusion to dynamically detect and replace uninformative tokens with projected tokens from other modalities for robust RGB-depth semantic segmentation, image-to-image translation, and 3D object detection. \cite{wang2023shaspec} proposed a model that learns modality-shared and modality-specific features for robust brain tumour segmentation. \cite{Choi2019embracenet} proposed a robust fusion strategy for multimodal classification. The main limitation of these methods is that they are generally designed for a specific modality combination and do not perform well when applied to other multimodal tasks \cite{lin2023vpfnet}. 

\textbf{Knowledge distillation and generation methods} have also become popular for robust MML. Studies by \cite{sharma2019gan1} and \cite{yu2018gan2} used GAN based generative models while \cite{dorent2019vae} used VAE based generative models for imputing missing modalities from available input modalities for unlerlying multimodal tasks. Recently \cite{ramazanova2024mmt} introduced an approach to learn missing modality tokens from available modalities. Different knowledge distillation approaches have also been applied in several multimodal tasks. \cite{tarvainen2017meanteacher} proposed mean teacher and \cite{maheshwari2023m3l} introduced multimodal teacher for semi-supervised image segmentation. \cite{shin2023crm} and \cite{hazarika2022robust_mmt2} applied self-distillation loss for robust RGB-thermal semantic segmentation. Apart from these approaches, weight space ensembling \cite{wortsman2022wiseft}, policy learning \cite{ma2022robust_mmt1}, optimal transport based approach \cite{han2022mmalign} and optimal fusion strategy designing \cite{maheshwari2023m3l} were also studied for robust MML for various tasks.

These approaches are either designed for specific tasks/modality combinations \cite{wang2022tf, shin2023crm, fan2023spidermesh} or require training extra modules/sub-networks \cite{tarvainen2017meanteacher, maheshwari2023m3l} for guiding the model under different missing modality scenarios. Our goal is to design a generic framework that is parameter-efficient and applicable to any model and modality combinations.
 
\textbf{Parameter-efficient network adaptation} has become very popular in recent years \cite{houlsby2019PEFT1, Liu2022PEFT2}. A number of parameter-efficient methods have been proposed for transfer learning \cite{he2022tl1, ding2023tl2} and uni-modal domain/task adaptation \cite{hu2021lora, lian2023ssf}. We can divide the approaches into following two major categories: \\
\textbf{Low-rank/additive adaptation} has been applied for uni-modal model fine-tuning and domain adaptation. For instance, LoRA \cite{hu2021lora}, QLoRA \cite{dettmers2023qlora}, KronA \cite{edalati2022krona} and KAdaptataion \cite{He2023kadaptation} learn low-rank factors for task/domain adaptation. Let $W$ be the weight matrix of any dense layer of a given pretrained uni-modal model. These approaches learn a low-rank weight update matrix $\Delta W$ to transform the input $x$ to that layer as
$h = Wx + \Delta Wx$,
where $h$ is the updated feature. Since the update matrix $\Delta W$ is low-rank, the number of learnable parameters remains a fraction of the total number of model parameters. \\
\textbf{Feature modulation based approach} is another parameter-efficient method to transform intermediate features of the pretrained model \cite{ioffe2015batchnorm, ba2016layernorm, wu2018group, lian2023ssf}. As shown in Figure~\ref{fig:intro-figure}c, it applies a linear transformation to the given input token/feature with learnable scale ($\gamma$) and shift ($\beta$) parameters. Given an input token $x$, this approach generates the output token as 
$h = \gamma \odot x + \beta$, 
where $\gamma,\beta,x,h$ are vectors of same dimension and $\odot$ represents element-wise multiplication along the embedding dimension. These scale ($\gamma$) and shift ($\beta$) parameters are input-independent %
and learned during the training process to help the model adjust and fine-tune its representations for better performance on the underlying task. 

Though parameter-efficient adaptation approaches have shown great potential in transfer learning, model fine tuning and task/domain adaptation, their potential remains unexplored in the context of missing modality in MML. In this study, we focus on parameter-efficient approaches to build our generic framework to enhance missing modality robustness in MML.

\section{Proposed Method}
\label{sec:proposed-methods}
In this section we first present a general framework for network adaptation for missing modalities. Then we discuss why we focus on parameter-efficient adaptation, present details of our proposed approach for missing modality adaptation and highlight the key benefits of our approach.

\subsection{Network Adaptation for Missing Modalities}
Let us denote the set of input modalities for a given multimodal task as $\mathcal{M} = \{m_1, \ldots, m_M\}$. Given the full set $\mathcal{M}$, we can train a model $f$ with parameters $\Theta_\mathcal{M}$ that maps inputs from all the  modalities (denoted as $\mathcal{X_M}$) to an output $y_\mathcal{M}$ as 
\begin{equation}
    y_\mathcal{M} = f(\mathcal{X_M}; \Theta_\mathcal{M}).
    \label{eq:all-modality-training}
\end{equation}

While we can ensure the availability of all input modalities during training, it is possible that some modalities may be inaccessible at test time, especially after real-world deployment. Any subset of modalities $\mathcal{M}$ can get missing due to hardware failure, data acquisition cost or privacy concerns. If we use a model trained on all the input modalities as denoted by \eqref{eq:all-modality-training}, significant performance drop is observed when a subset of modalities gets missing during test time as shown in Table~\ref{tab:missing-modality-effect-on-cmnext}. 

\subsubsection{Na\"ive approach}
When a subset of the modalities $\mathcal{M}$ is missing, a simple and na\"ive approach is to train a new model for the available input modalities. Without loss of generality, suppose $\mathcal{K} \subset \mathcal{M}$ represents missing modalities. We can use the available input modalities $\mathcal{S} = \mathcal{M} \setminus \mathcal{K}$ to retrain the model $f$ for a new set of parameters $\Theta_\mathcal{S}$ as
\begin{equation}
    y_{\mathcal{S}} = f(\mathcal{X_S}; \Theta_\mathcal{S}), 
    \label{eq:missing-modality-training}
\end{equation}
where $\mathcal{X_S}$ represents input data for modalities in $\mathcal{S}$.  
In principle, we can train one model for every possible $\mathcal{S}\subset \mathcal{M}$ and use the corresponding model at the test time. Such an approach is infeasible because of computational and storage resources required to train models for a large number of possible modality combinations. Furthermore, deploying a large number of trained models and selecting one of them at test time is not feasible in real-world scenarios. Another drawback of this method is that, even though we would like $y_\mathcal{S} \approx y_\mathcal{M}$, the training process mentioned earlier does not guarantee it.

\subsubsection{Parameter-efficient approach}
We propose an alternative approach to adapt a single model for all subsets of input modalities $\mathcal{S}\subset \mathcal{M}$ in a parameter-efficient manner. 
First, we select a model $f$ trained on the full set of modalities $\mathcal{M}$ as shown in \eqref{eq:all-modality-training} and freeze the parameters $\Theta_\mathcal{M}$. Then we learn a small number of parameters $\Delta_\mathcal{S}$, specific to the available input modality set $\mathcal{S}$, and update the model as %
\begin{equation}
    \hat{y}_{\mathcal{S}} = f(\mathcal{X_S}; \Theta_\mathcal{M}, \Delta_\mathcal{S}), 
    \label{eq:model-adaptation-training}
\end{equation}
where $\hat{y}_\mathcal{S}$ represents the prediction of the updated model. Our goal is to keep $\hat{y}_\mathcal{S}$ close to all modality prediction ${y}_\mathcal{M}$ in the best case ($\hat{y}_\mathcal{S} \approx {y}_\mathcal{M}$) and close to the prediction ${y}_\mathcal{S}$ made by a model trained on the available input modalities in the worst case ($\hat{y}_\mathcal{S} \approx {y}_\mathcal{S}$).

The adaptation method shown in \eqref{eq:model-adaptation-training} is considered parameter-efficient if the number of parameters in $\Delta_\mathcal{S}$ is significantly smaller compared to the total number of parameters in $\Theta_\mathcal{M}$. During adaptation, we keep $\Theta_\mathcal{M}$ frozen and demonstrate that less than $1\%$ of the total parameters for $\Delta_\mathcal{S}$ are sufficient for network adaptation (Section~\ref{sec:performance-vs-params}).

\subsubsection{Need for parameter-efficient adaptation}
In recent years, a number of approaches have been proposed for MML with missing modalities. To the best of our knowledge, parameter-efficient adaptation is still unexplored in this field. The current methods for robust MML, as discussed in Section~\ref{sec:related-works}, require retraining the whole model with specialized training strategy \cite{shin2023crm, hussen2020mdrop} or utilize extra module/sub-network to guide the multi-modal model \cite{maheshwari2023m3l, tarvainen2017meanteacher}. Furthermore, these methods are not very generic and do not perform well on different missing modality scenarios as shown in Table~\ref{tab:MFNet-comparison-with-other-models} and \ref{tab:NYU-comparison-with-other-models}. To solve these issues, we propose parameter-efficient adaptation for enhancing missing modality robustness of MML. Our approach requires learning a very small number of parameters for different missing modality scenarios without the need to retrain the whole network. Furthermore, it is also applicable to diverse model architectures, tasks and modality combinations as discussed in Section~\ref{sec:results}.

\subsection{Parameter-Efficient Adaptation for Robust MML}
This section outlines our approach for multimodal network adaptation for missing modalities. We explain the reasons behind selecting intermediate feature modulation and compare with other parameter-efficient methods, highlighting key benefits of our approach.

{\bf Adaptation for multimodal models.} To the best of our knowledge, no parameter-efficient adaptation approach has been proposed or applied for multimodal model adaptation to handle missing modalities. 
We draw our motivation from low-rank adaptation \cite{hu2021lora, zaken2022bitfit, ba2016layernorm, ioffe2015batchnorm} and feature modulation based approach \cite{lian2023ssf}. 

These approaches can enhance the representation capabilities of deep models. We extend these adaptation approaches to build a generic framework that can transform the intermediate features of the available modalities to find an optimal feature representation to compensate for the performance gap due to missing modalities.

\subsubsection{Training: model adaptation for missing modalities}

Our approach is illustrated in Figure~\ref{fig:intro-figure}. Without loss of generality, let us assume a generic multi-modal model in which each modality goes through a separate encoder for feature extraction, followed by a fusion block to fuse the extracted features. The fused feature is passed to a decoder head for making prediction. This setup can be easily generalized to models with shared encoder, different encoder/model architecture and/or different (early or mid) fusion strategy.

We train this multimodal network $f$ with all available modalities in $\mathcal{M}$ to learn the parameters $\Theta_\mathcal{M}$ as shown in \eqref{eq:all-modality-training}. Then we adapt $f$ for different subsets of available modalities $\mathcal{S}\subset \mathcal{M}$. Unlike existing methods, we do not try to generate \cite{ma2021smil, bachmann2022multimae}, approximate \cite{wang2022tf, wang2023shaspec} or distill knowledge \cite{shin2023crm, maheshwari2023m3l} from any other modality/sub-network. 
Our goal is to learn a modified function for the available input modalities to appropriately learn and fuse features to compensate for any missing modality. Instead of re-training the entire network on the available modalities as shown in \eqref{eq:missing-modality-training}, we adapt the base network $f$ and focus on learning a minimal set of parameters following \eqref{eq:model-adaptation-training}.

To adapt the base model $f$, as shown is Figure~\ref{fig:intro-figure}a, we freeze the parameters $\Theta_\mathcal{M}$ (marked as {\color{snowflakeColor}\faSnowflake} in light blue rectangles), which freezes all the layers in the model. Then we insert adaptable layers with learnable parameters $\Delta_\mathcal{S}$ (marked as {\color{fireColor}\faFire} in light orange rectangles) after each frozen linear, convolutional, and norm layers.
We show the missing modality branches as grayed-out indicating that they are inactive and do not contribute to the model output. Then we adapt $f$ following \eqref{eq:model-adaptation-training} to learn $\Delta_\mathcal{S}$. While learning $\Delta_\mathcal{S}$ for a given modality combination, $\mathcal{S}$, we set the missing modalities to zero following standard practice \cite{hussen2020mdrop, maheshwari2023m3l, shin2023crm, lin2023vpfnet}.  We minimize the cross-entropy loss with respect to $\Delta_\mathcal{S}$ for different modality combinations. 

Below we discuss how to use low-rank  and intermediate feature modulation-based multimodal network adaptation to accommodate missing modalities. Our framework is generic and can also incorporate other parameter-efficient adaptation approaches.

\noindent \textbf{Low-rank/additive adaptation.} We extend low-rank/additive approaches to adapt multimodal model for missing modalities. Let us assume that $W_m$ be one of the weight matrices from any layer for the $m^{th}$ input modality where $m \in \mathcal{S}$. As shown in Figure~\ref{fig:intro-figure}b, we learn a low-rank weight update matrix $\Delta W_m$ for that layer to transform the input $h_{m,i}$ to the layer as
\begin{equation}
    h_{m,o} = W_mh_{m,i} + \Delta W_mh_{m,i},\text{ for all } m \in \mathcal{S},
    \label{eq:mm-lora}
\end{equation}
where $h_{m,o}$ is the transformed output feature that is passed to the next layer in the model. Since $\Delta W_m$ is low-rank, the total number of learnable parameters remains a fraction of the total number of model parameters. We can represent the learnable parameters $\Delta_\mathcal{S} = \{\Delta W_m \}_{m \in \mathcal{S}}$ as the collection of all low-rank update matrices. 

\noindent \textbf{Intermediate feature modulation.} We extend SSF \cite{lian2023ssf} method to work with multimodal models with missing modalities. The adaptable SSF layers modulate the intermediate tokens/features from each available modality at every layer as shown in Figure~\ref{fig:intro-figure}c. 
For the $m^{th}$ input modality where $m \in \mathcal{S}$, we denote the learnable scale and shift parameters as $\gamma_m \in \mathbb{R}^{d}$ and $\beta_m \in \mathbb{R}^{d}$ respectively where $d$ is the embedding dimension of the model. The output ${h}_{m,o} \in \mathbb{R}^{N \times d}$ form any frozen layer for the $m^{th}$ input modality goes through the SSF layer that follows it. The SSF layer applies a linear transformation on ${h}_{m,o}$ as follows:
\begin{equation}
    {h}_{m,i} = \gamma_m \odot {h}_{m,o} + \beta_m, \text{ for all } m \in \mathcal{S}, 
    \label{eq:mm-ssf-eqn}
\end{equation}
where ${h}_{m,i} \in \mathbb{R}^{N \times d}$ is the transformed feature which is fed to the next frozen layer in the model and $N$ is the number of tokens. Note that if the output of any layer is of shape $(H \times W \times d)$ (for convolutional layers), we reshape it to $(N \times d)$, where $N = H \times W$, before applying \eqref{eq:mm-ssf-eqn}. We reshape the transformed feature back to the original shape (if required) before passing it to the next layer. We can represent the learnable parameters as $\Delta_\mathcal{S} = \{\gamma_\mathcal{S}, \beta_\mathcal{S}\} = \{\gamma_m, \beta_m\}_ {m \in \mathcal{S}}$. 
BitFit \cite{zaken2022bitfit} method can also be used for adaptation as we only need to learn the bias/shift terms $\beta_m$ for all $m \in \mathcal{S}$. We modify \eqref{eq:mm-ssf-eqn} as
\begin{equation}
    {h}_{m,i} = {h}_{m,o} + \beta_m, \text{ for all }m \in \mathcal{S}, 
    \label{eq:mm-bitfit-eqn}
\end{equation}
and the learnable parameters can be represented as $\Delta_\mathcal{S} = \{\beta_m\}_{m \in \mathcal{S}}$. Thus the intermediate features from each available modality are modulated to find a better representation to compensate for the missing modalities.

\subsubsection{Inference: model adaptation for missing modalities}

At the test time, we load the base multimodal model $f$ with the pretrained weights $\Theta_\mathcal{M}$. If all the modalities are available, then we can use $\Theta_\mathcal{M}$ to make predictions. 
When a subset of the modalities are missing, we can select the learned parameters $\Delta_\mathcal{S}$ corresponding to the available input modalities $\mathcal{S}$, insert them into the model and use them to make prediction as follows: 
\begin{equation}
    \hat{y}_{\mathcal{S}} = 
    \begin{cases}
        f(\mathcal{X_S}; \Theta_\mathcal{M}) & \text{if } \mathcal{S} = \mathcal{M},\\
        f(\mathcal{X_S}; \Theta_\mathcal{M}, \Delta_\mathcal{S}) & \text{if } \mathcal{S} \subset \mathcal{M}.
    \end{cases}
\end{equation}

Since we are inserting the adaptable layers after each layer, it does not require any major change to the model architecture and can be done easily without reloading all the model parameters $\Theta_\mathcal{M}$. We just need to load the parameters in $\Delta_\mathcal{S}$ and insert them into the model. Since we only insert a very small number of additional parameters, it adds very limited computational overhead. Furthermore, if a different subset of modalities becomes available, the adjustment is straightforward. We only need to replace the existing learned parameters $\Delta_\mathcal{S}$ with the corresponding parameters for the available modality set, ensuring adaptability and flexibility in handling diverse combinations of available modalities during the testing phase.

We only insert adaptable layers in the encoders and fusion blocks, while keeping the decoder/prediction head unchanged. We observed that using pretrained decoder/prediction head provided a good overall performance with several missing modalities.

\subsubsection{Feature modulation vs low-rank adaptation}
\label{sec:ssf-vs-lora}

While we present three adaptation approaches in \eqref{eq:mm-lora}, \eqref{eq:mm-ssf-eqn}, and \eqref{eq:mm-bitfit-eqn}, we select intermediate feature modulation with SSF \eqref{eq:mm-ssf-eqn} as the main approach for our experiments. 
We primarily selected this technique because of its simplicity and effectiveness. Our experiments show that feature transformation via simple linear transformation with SSF works well for most of the scenarios compared to other parameter-efficient adaptation approaches as summarized in Table~\ref{tab:overall-comparison-with-other-methods}. We provided a detailed comparison in terms of mean accuracy, F1 score and \% mIoU in Table~\ref{tab:MFNet-comparison-with-other-methods}, \ref{tab:nyu-other-methods} and \ref{tab:mcubes-other-methods} in the supplementary section. 
SSF shows great promise in enhancing representation power \cite{ba2016layernorm}, faster convergence \cite{ioffe2015batchnorm}, prevents loss of information in the representation learning process \cite{wu2018group} and mitigates distribution mismatch between the upstream and downstream tasks \cite{lian2023ssf}. These characteristics motivated us to extend this method for MML with missing modalities and build a generic framework that is very effective in learning the proper modulation of available input modalities to bridge the performance gap in the face of missing modalities.

{\bf Some key benefits} of this approach are as follows. First, The parameters \{$\gamma,\beta$\} are independent of the input features/modalities, which makes it applicable to diverse tasks and input modality combinations. Second, we can easily insert these learnable layers in existing models without changing the model architecture. We can easily switch/select the corresponding SSF parameters for a given input modality combination. Finally, it introduces extremely small number of additional learnable parameters. The resulting adaptation offers significant savings compared to training a separate model for each input combination or retraining the model using some specialized training strategy like modality dropout \cite{hussen2020mdrop, neverova2015moddrop} or knowledge distillation \cite{shin2023crm, maheshwari2023m3l}.

\section{Experiments and Results}
\label{sec:results}

We performed detailed experiments to evaluate the effectiveness and generalizability of our proposed method on five multimodal tasks across seven datasets. In this section, we present comparison with existing baseline methods that are robust to missing modalities. 

\subsection{Datasets and Tasks}
\label{sec:datasets-and-tasks} 
In this section, we provide a brief description of each dataset. Please refer to Section~\ref{supp-datasets} in the supplementary materials for comprehensive details on each dataset.

\subsubsection{Multimodal semantic segmentation.}
\textbf{MFNet dataset} \cite{Ha2017mfnet} contains 1569 registered RGB-Thermal image pairs and divided into train and test sets. Each image is $640 \times 480$ pixels, contains annotation for 9 classes.\\
\textbf{NYUDv2 dataset} \cite{Silberman2012nyudv2} has 1449 pairs of aligned RGB-Depth image pairs. It is divided into train and test sets having 795 and 654 image pairs respectively. Each image is $640 \times 480$ pixels and contains annotation for 40 classes. We used HHA encoded images \cite{gupta2014hha} instead of raw depth maps for our experiments.

\subsubsection{Multimodal material segmentation}
\textbf{MCubeS dataset} \cite{Liang2022mcubes} has 4 input modalities: RGB, Angle of Linear Polarization (AoLP), Degree of Linear Polarization (DoLP) and Near-Infrared (NIR). The dataset is divided into train, validation and test sets containing 302, 96 and 102 sets of images respectively along with ground truth per-pixel annotation for 20 material classes.

\subsubsection{Multimodal action recognition}
\textbf{NTU RGB+D (NTU60) dataset} \cite{shahroudy2016ntudataset} contains 56,880 video samples across 60 action classes. It contains RGB videos $(1920 \times 1080)$, depth map sequences $(512 \times 424)$, infrared (IR) videos $(512 \times 424)$ and 3D skeletal data. We use RGB and depth data for our experiments and evaluate performance using cross subject protocol.

\subsubsection{Multimodal sentiment analysis}
\textbf{CMU-MOSI dataset} \mbox{\cite{zadeh2016mosi}} contains audio, visual and text modality for multimodal sentiment analysis. The dataset has 2,199 samples divided into train, validation and test containing 1,284, 229 and 686 samples respectively.\\
\textbf{CMU-MOSEI dataset} \mbox{\cite{bagher2018mosei}} is another large scale dataset. It contains 23,453 samples having audio, visual and text. The dataset is divided into train, validation and test sets.

\subsubsection{Multimodal classification}
\textbf{UPMC Food-101 dataset} \cite{wang2015food101} is a popular multimodal classification dataset containing image and text as input modalities. The dataset contains 90,704 image-text pairs and 101 food categories. 

\subsection{Implementation Details}
\label{sec:impl-details}
We use CMNeXt \cite{zhang2023cmnext} as the base model for multimodal segmentation tasks, multimodal transformer  \cite{tsai2019mult} for multimodal sentiment analysis, UMDR \cite{zhou2023umdr} for multimodal action recognition and ViLT \cite{kim2021vilt} for multimodal classification. We train the corresponding base model with all the input modalities for each dataset. To evaluate performance with missing modalities, we provide the available modalities and set the missing modalities to zero for images and empty string for texts. To perform model adaptation for any modality subset $\mathcal{S} \subset \mathcal{M}$, we fine tune the learnable parameters until convergence for all the tasks. 

For multimodal segmentation tasks, we set the learning rate to $6\times 10^{-5}$ and apply polynomial learning rate scheduler with power = 0.9. The first 10 epochs are warm-up epochs and the learning rate is set to 0.1 times the original rate. The scale $(\gamma)$ and shift $(\beta)$parameters were initialized with all 1s and 0s respectively. We use cross-entropy loss and AdamW optimizer \cite{loshchilov2017adamw}, with $\epsilon = 10^{-8}$ and weight decay = 0.01. We set batch size to 4 and report single scale performance. All other hyper-parameters are the same as \cite{zhang2023cmnext}. For multimodal sentiment analysis, action recognition and classification tasks, we used the default settings from \cite{yu2021mmsa-code}, \cite{zhou2023umdr} and \cite{lee2023map} respectively. Please refer to Section~\ref{sec:impl-details-supplementary} in the supplementary materials for additional details.

For every task/dataset, we show the reported results from prior works where possible. It is important to note that, because of this criteria, some of the baseline methods may only be present in specific experiments depending on the availability of their reported numbers. We also perform detailed comparison of SSF with other parameter-efficient adaptation techniques which we discuss in Section~\ref{sec:other-peft-comparison} in the supplementary materials.

\begin{table}[t]
    \centering
    \setlength{\tabcolsep}{2pt}
    \caption{Performance comparison with different baseline methods for multimodal semantic segmentation on MFNet and NYUDv2 datasets and multimodal material segmentation on MCubeS dataset. We use CMNeXt as the base model. {\textbf{Bold}} letters represent best results.}
    \begin{tabular}{cllcccc}
        \toprule
        Dataset & \multicolumn{1}{c}{Input}         & \multicolumn{1}{c}{Missing} & {Pretrained}   & {Modality Duplication} & {Dedicated}  & { Adapted (Ours)} \\
        \midrule
        \midrule
        \multirow{3}{*}{MFNet}  & RGB-Thermal       & -                           & 60.10     & 60.10 & 60.10          & 60.10              \\
                                & RGB               & Thermal                     & 53.71     & 52.33 & \textbf{55.86} & 55.22          \\
                                & Thermal           & RGB                         & 35.48     & 44.43 & \textbf{53.34} & 50.89          \\
        \midrule
        \multirow{3}{*}{NYUDv2} & RGB-Depth         & -                           & 56.30     & 56.30 & 56.30          & 56.30              \\ 
                                & RGB               & Depth                       & 51.19     & 46.19 & 52.18          & \textbf{52.82} \\
                                & Depth             & RGB                         & 5.26      & 13.94 & 33.49          & \textbf{36.72} \\ 
        \midrule
        \multirow{4}{*}{MCubeS} & RGB-AoLP-DoLP-NIR & -                           & 51.54     & 51.54 & 51.54          & 51.54              \\
                                & RGB-AoLP-DoLP     & NIR                         & 49.06     & 49.93 & 49.48          & \textbf{51.11} \\
                                & RGB-AoLP          & DoLP-NIR                    & 48.81     & 49.23 & 48.39          & \textbf{50.62} \\
                                & RGB               & AoLP-DoLP-NIR               & 42.32     & 48.96 & 48.11          & \textbf{50.43} \\
        \bottomrule
    \end{tabular}
    \label{tab:missing-modality-effect-on-cmnext}
\end{table}

\subsection{Experiments on Multimodal Segmentation}
\label{sec:result-segmentation}
In this section, we present experimental results for multimodal semantic and material segmentation. First, we show an overall comparison of our approach with baselines methods and then we compare with existing robust methods.

\subsubsection{Overall performance comparison}
We report experimental results for different baseline methods in Table~\ref{tab:missing-modality-effect-on-cmnext}. 
\textbf{Pretrained} model refers to the base CMNeXt model trained with all the available modalities. 
\textbf{Modality Duplication} means that one of the available modalities is used as a substitution for the missing modality.
\textbf{Dedicated} training indicates that we train one CMNeXt model for each input modality combination and use the model corresponding to the available modalities when some modalities get missing.
\textbf{Adapted} model refers to the model that is adapted using our approach for each input modality combination. 

Pretrained model show significant performance drop with missing modalities. We see a 6.39\% and 5.11\% drop when Thermal is missing on MFNet and Depth is missing on NYUDv2, respectively, compared to the case when all modalities are available. The effect is amplified when RGB gets missing as we observe 24.62\% and 51.04\% drop on MFNet and NYUDv2 dataset respectively. On MCubeS dataset, we observe 2.48--9.22\% drop in pretrained model when different modality combinations are missing. Similar trend of performance drop is observed for modality duplication approach though it performs better than pretrained models for most of the cases.

The overall performance of the Adapted model is significantly better than Pretrained model and Modality Duplication approach. For MFNet, an improvement of 1.51\% and 15.41\% is observed compared to the Pretrained model when RGB and Thermal are available respectively. The performance of the Adapted models is also close to the Dedicated models. For NYUDv2 dataset, we see 1.63\% and 31.46\% performance improvement compared to Pretrained model when depth and RGB are missing, respectively. For all input combinations on MCubeS dataset, we see 1.82--8.11\% performance improvement compared to the Pretrained model. The Adapted model performs better than Dedicated models on NYUDv2 and MCubeS datasets. Per-class IoU analysis shows that adapted models perform better than pretrained models for most of the classes which provides an overall performance improvement as discussed in Section~\ref{sec:per-class-iou-supp}.

Feature modulation during adaptation helps the model learn better feature representation and thus it performs better when modalities are missing. We discuss this in Section~\ref{sec:mcubes-cosine-sim}. Results also indicates that we do not need to train a dedicated network for each modality combination which requires more time and computation resources. Rather adapting one base model is sufficient to have comparable or even better performance in missing modality scenarios with less time and computational overhead.

\subsubsection{Comparison with robust methods on MFNet dataset}
We compare the performance of the Adapted model with existing robust models for RGB-thermal semantic segmentation on MFNet dataset in Table~\ref{tab:MFNet-comparison-with-other-models}. Results show that the Adapted model offers the best average performance compared to existing baseline methods. Among the robust models, complementary random masking and knowledge distillation based model CRM \cite{shin2023crm} shows competitive performance with the Adapted model. The Adapted model performs better when only RGB is available while CRM performs better when only Thermal is available. Notably CRM is designed specifically for RGB-Thermal pairs and requires specialized training approach. In contrast, our approach is generic, applicable to any input modalities and does not require any specialized training technique. Our approach performs significantly better compared to partial masking and recursive meshing based SpiderMesh \cite{fan2023spidermesh}, variational pobabilistic fusion based VPFNet \cite{lin2023vpfnet} and modality discrepancy reduction based MDRNet \cite{zhao2023mdrnet} models.

\begin{table}[t]
    \centering
    \caption{Performance comparison with existing robust methods for MFNet dataset. RGB and Thermal columns report performance when only RGB and only Thermal are available. Average column reports average performance when one of the two modalities gets missing. `-' indicates that results for those cells are not published. $^\ast$ indicates that available code and pretrained models from the authors were used to generate the results.}
    \setlength{\tabcolsep}{4pt}
    \centering
    \begin{tabular}{llccccccc}
        \toprule
        \multicolumn{1}{c}{\multirow{2}{*}{Methods}} & \multicolumn{1}{c}{\multirow{2}{*}{Backbone}} & \multicolumn{1}{c}{\multirow{2}{*}{Parameters (M)}} & \multicolumn{2}{c}{RGB} & \multicolumn{2}{c}{Thermal} & \multicolumn{2}{c}{Average} \\ 
        \multicolumn{3}{c}{}                & mAcc       & \% mIoU           & mAcc       & \% mIoU           & mAcc       & \% mIoU \\ 
        \midrule
        \midrule
        FuseNet \cite{haz2016fusenet}       & VGG-16 \cite{karen2015vgg16} & - & 11.11          & 10.31          & 41.33          & 36.85          & 26.22      & 23.58 \\
        MFNet \cite{Ha2017mfnet}            & DCNN \cite{bengio2019DCNN} & 0.73 & 26.62          & 24.78          & 19.65          & 16.64          & 23.14      & 20.71 \\
        RTFNet \cite{sun2019rtfnet}         & ResNet-152 \cite{he2016resnet} & 254.51 & 44.89          & 37.30          & 26.41          & 24.57          & 35.65      & 30.94 \\
        SAGate \cite{chen2020SAGate}        & ResNet-50 \cite{he2016resnet} & 110.85 & 32.01          & 30.57          & 13.34          & 12.51          & 22.68      & 21.54 \\
        FEANet \cite{deng2021feanet}        & ResNet \cite{he2016resnet} & - & 15.96          & 8.69           & 58.35          & 48.72          & 37.16      & 28.71 \\ 
        MDRNet \cite{zhao2023mdrnet}        & ResNet-50 \cite{he2016resnet} & 64.60 & 57.11          & 45.89          & 41.98          & 30.19          & 49.55      & 38.04 \\
        VPFNet \cite{lin2023vpfnet}         & ResNet-50 \cite{he2016resnet} & - & 48.14          & 41.08          & 42.20          & 35.80          & 45.17      & 38.44 \\
        SpiderMesh \cite{fan2023spidermesh} & ResNet-152 \cite{he2016resnet} & 151.81 & -              & 39.60          & -              & 50.50          & -          & 45.05 \\
        CRM \cite{shin2023crm}              & Swin-S \cite{liu2021swin} & 117.68 & -              & 52.70          & -              & \textbf{53.10}          & -      & 52.90 \\
        CMNeXt \cite{zhang2023cmnext}$^\ast$             & MiT-B4 \cite{xie2021segformer} & 116.56 & 60.74              & 53.71          & 38.18              & 35.48          & 49.46      & 44.60 \\
        \textbf{Adapted (Ours)}             & MiT-B4 \cite{xie2021segformer} & 117.35 &  \textbf{67.18} & \textbf{55.22}           & \textbf{66.70}          & 50.89          & \textbf{66.94} & \textbf{53.06} \\
        \bottomrule
    \end{tabular}
    \label{tab:MFNet-comparison-with-other-models}
\end{table}

\begin{table}[t]
    \centering
    \caption{Performance comparison with existing robust methods for NYUDv2 dataset. RGB and Depth columns report performance when only RGB and only Depth are available. Average column indicates average performance when one of the two modalities gets missing.  $^\ast$ indicates that available code and pretrained models from the authors were used to generate the results. Other results are from the corresponding papers.}
    \setlength{\tabcolsep}{4pt}
    \centering
    \begin{tabular}{llccccccc}
        \toprule
        \multicolumn{1}{c}{\multirow{2}{*}{Methods}} & \multicolumn{1}{c}{\multirow{2}{*}{Backbone}} & \multicolumn{1}{c}{\multirow{2}{*}{Parameters (M)}} & \multicolumn{2}{c}{RGB} & \multicolumn{2}{c}{Depth} & \multicolumn{2}{c}{Average} \\ 
        \multicolumn{1}{c}{}                         & & & mAcc        & \% mIoU      & mAcc           & \% mIoU           & mAcc           & \% mIoU \\ 
        \midrule
        \midrule
        FCN \cite{long2015fcn}                       & VGG-16 \cite{karen2015vgg16} & 134.00 & 44.70       & 31.60     & 35.70          & 25.20          & 40.20          & 28.40 \\
        Dilated FCN-2s \cite{Kamran2018eyd}          & VGG-19 \cite{karen2015vgg16} & 55.81 & 47.10       & 32.30     & 39.30          & 26.80          & 43.20          & 29.55 \\
        AsymFusion \cite{wang2020asymfusion} & ResNet-101 \cite{he2016resnet} & 118.20 & 59.00       & 46.50     & 45.60          & 34.30          & 52.30          & 40.40 \\
        CEN \cite{wang2020cen}$^\ast$        & ResNet-101 \cite{he2016resnet} & 118.20 & 51.77       & 39.59     & 28.98          & 19.32          & 40.38          & 29.46 \\
        TokenFusion \cite{wang2022tf}$^\ast$         & MiT-B3 \cite{xie2021segformer} & 45.92 & 63.49       & 49.32     & 46.83          & \textbf{36.84} & 55.16          & 43.08 \\
        CMNeXt \cite{zhang2023cmnext}$^\ast$         & MiT-B4 \cite{xie2021segformer} & 116.56 & 64.10       & 51.19     & 8.30          & 5.26 & 36.20          & 28.23 \\
        \textbf{Adapted (Ours)}        & MiT-B4 \cite{xie2021segformer} & 117.35 & \textbf{67.96} & \textbf{52.82} & \textbf{52.42} & 36.72  & \textbf{60.19} & \textbf{44.77} \\
        \bottomrule
    \end{tabular}
    \label{tab:NYU-comparison-with-other-models}
\end{table}

\begin{figure}[t]
    \centering
    \subfloat[Visualization of predictions on MFNet dataset for multimodal semantic segmentation\label{fig:vis-mfnet-1}]{\includegraphics[width=0.95\textwidth]{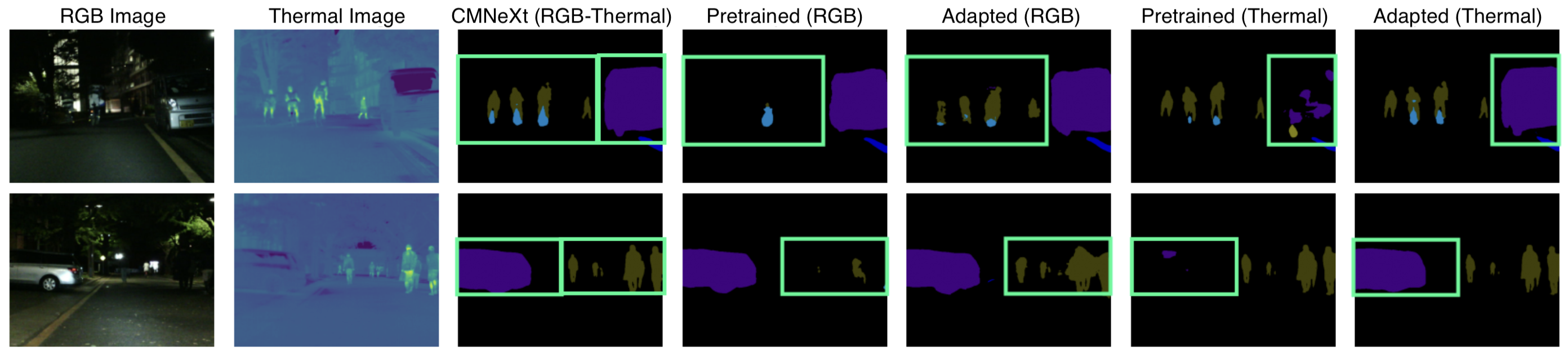}}
    
    \hfill
    \\[3.5pt]
    
    \subfloat[Visualization of predictions on NYUDv2 dataset for multimodal semantic segmentation\label{fig:vis-nyu-1}]{\includegraphics[width=0.95\textwidth]{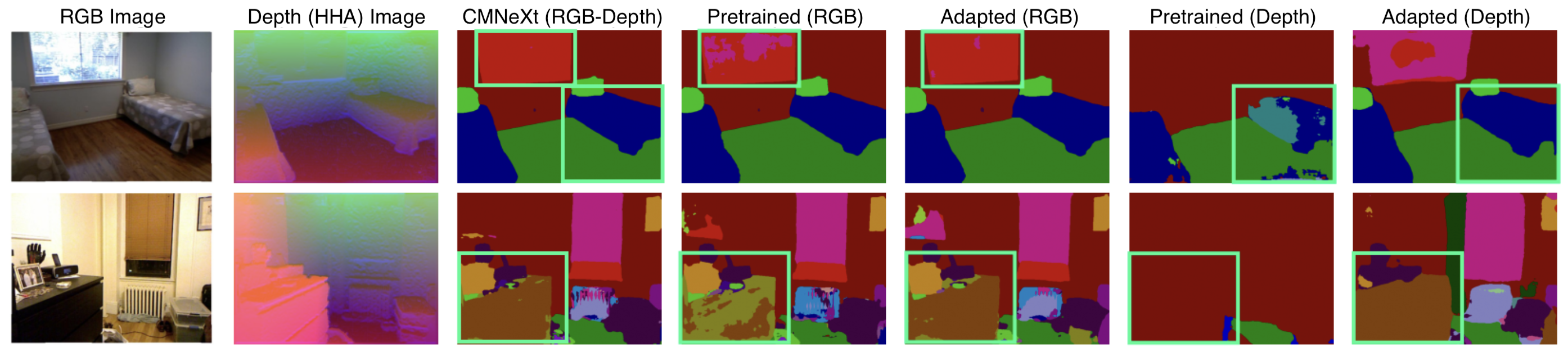}}
    
    \hfill
    \\[3.5pt]
    \subfloat[Visualization of predictions on MCubeS dataset for multimodal material segmentation\label{fig:vis-mcubes-1}]{\includegraphics[width=0.95\textwidth]{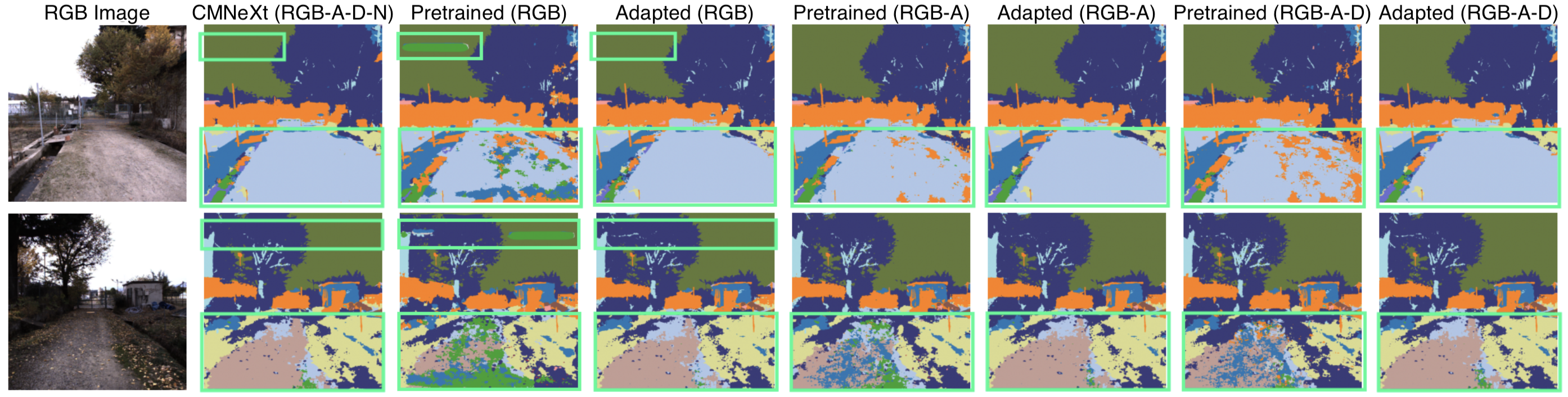}}
    
  \caption{Examples of predicted segmentation maps for the Pretrained and Adapted models. Title above each subimage shows method name (available modalities). CMNeXt column shows the predictions with all the modalities. Segmentation quality improves significantly after model adaptation for all input modality combinations. Green boxes highlight areas with salient differences in results (e.g., cars and humans missing in the Pretrained model with missing modalities but visible in the Adapted model). For MCubeS dataset, we only show RGB input images for brevity. A, D and N denote angle of linear polarization, degree of linear polarization, and near-infrared, respectively. }
  \label{fig:vis-prediction}
\end{figure}

\subsubsection{Comparison with robust methods on NYUDv2 dataset}
Table~\ref{tab:NYU-comparison-with-other-models} shows the performance comparison with existing robust models for RGB-Depth semantic segmentation on NYUDv2 dataset. On an average, the Adapted model performs better than the existing robust models. Dynamic token selection and substitution based model TokenFusion \cite{wang2022tf} performs slightly better (+0.12\%) in mIoU when Depth is available and RGB is missing, but shows larger drop (-5.59\%) in mean accuracy. On the other hand, the Adapted model performs significantly better (+3.5\% mIoU and +4.47\% mean accuracy) when RGB is available and Depth is missing. The average performance of the Adapted model is also better than the TokenFusion model despite the fact that TokenFusion was designed to work with RGB-Depth pair, whereas our approach is independent of input modalities. Our method also performs significantly better compared to dynamic channel exchange based CEN \cite{wang2020cen} and asymmetric fusion based AsymFusion \cite{wang2020asymfusion} models. 

We observe that the CMNeXt model performs poorly when Depth is available and RGB is missing. This is due to its asymmetric architecture, which treats RGB as the primary modality and others as supplementary. As a result, performance drops significantly in the absence of RGB. However, the model overcomes this issue after adaptation and improves performance in all missing modality scenarios demonstrating the effectiveness of our adaptation approach. 

\subsubsection{Visualization of predictions}
For qualitative analysis, we show some examples of the predicted segmentation maps form the Pretrained and Adapted models in Figure~\ref{fig:vis-prediction}. For each dataset, we show the input images, predictions when all the modalities are available (CMNeXt column), predictions from the pretrained and adapted models for different available/missing modality scenarios (Available input modality names are shown in parentheses above each image). We see in Figure~\ref{fig:vis-mfnet-1}, the Pretrained model fails to detect humans when only RGB images are available and cars when only Thermal images are available. The adapted model can detect both humans and cars with missing modalities. 

On NYUDv2 dataset, as shown in Figure~\ref{fig:vis-nyu-1}, the Adapted model can detect window, bed, and furniture with higher accuracy than the Pretrained model with missing modalities. On MCubeS dataset, the Adapted model can identify sand, sky, and gravel with higher accuracy than the pretrained model. In all cases, the predictions from the Adapted model with missing modalities are closer to the predictions of the pretrained model with all the input modalities. We provide additional visualizations in Figure~\ref{fig:vis-prediction-more} in the supplementary materials.

\begin{table}[t]
    \centering
    \setlength{\tabcolsep}{3pt}
    \caption{Comparison of our adaptation technique with existing methods for multimodal sentiment analysis on CMU-MOSI and CMU-MOSEI datasets.}
    \resizebox{\textwidth}{!}{
        \begin{tabular}{lllccccccccc}
            \toprule
            \multicolumn{1}{c}{\multirow{2}{*}{Datasets}} &
              \multicolumn{1}{c}{\multirow{2}{*}{Methods}} &
              \multicolumn{1}{c}{\multirow{2}{*}{Backbone}} &
              \multicolumn{1}{c}{\multirow{2}{*}{Parameters (M)}} &
              \multicolumn{2}{c}{Audio} &
              \multicolumn{2}{c}{Visual} &
              \multicolumn{2}{c}{Audio-Visual} &
              \multicolumn{2}{c}{Average} \\
            \multicolumn{1}{c}{} &
            \multicolumn{1}{c}{} &
            \multicolumn{1}{c}{} &
              \multicolumn{1}{c}{} &
              ACC &
              F1 &
              ACC &
              F1 &
              ACC &
              F1 &
              ACC &
              F1 \\
            \midrule
            \midrule
            \multirow{6}{*}{CMU-MOSI} &
              MulT \cite{tsai2019mult} &
              Transformer \cite{vaswani2017attention} &
              2.58 &
              48.31 &
              40.98 &
              52.44 &
              51.77 &
              48.93 &
              41.95 &
              49.89 &
              44.90 \\
             &
              MFN \cite{Zadeh_2018_MFN} &
              LSTM \cite{hochreiter1997lstm} &
              2.17 &
              56.86 &
              44.81 &
              55.95 &
              42.94 &
              56.86 &
              51.07 &
              56.56 &
              46.27 \\
             &
              TFN \cite{zadeh2017_TFN} &
              LSTM \cite{hochreiter1997lstm} &
              5.04 &
              42.23 &
              25.07 &
              42.38 &
              25.40 &
              42.23 &
              25.07 &
              42.28 &
              25.18 \\
             &
              BERT\_MAG \cite{rahman2020_bertmag} &
              BERT \cite{kenton2019bert} &
              110.83 &
              \textbf{57.77} &
              42.31 &
              \textbf{57.77} &
              42.31 &
              \textbf{57.77} &
              42.31 &
              \textbf{57.77} &
              42.31 \\
             &
              LMF \cite{Liu2018_LMF} &
              LSTM \cite{hochreiter1997lstm} &
              1.10 &
              42.23 &
              25.07 &
              43.14 &
              27.54 &
              43.29 &
              27.61 &
              42.89 &
              26.74 \\
             &
              \textbf{Adapted (Ours)} &
              Transformer \cite{vaswani2017attention} &
              2.60 &
              50.00 &
              \textbf{46.71} &
              54.88 &
              \textbf{54.39} &
              55.49 &
              \textbf{53.96} &
              53.46 &
              \textbf{51.69} \\
            \midrule
            \multirow{6}{*}{CMU-MOSEI} &
              MulT \cite{tsai2019mult} &
              Transformer \cite{vaswani2017attention} &
              2.58 &
              37.15 &
              20.12 &
              38.28 &
              23.70 &
              41.91 &
              32.78 &
              39.11 &
              25.53 \\
             &
              MFN \cite{Zadeh_2018_MFN} &
              LSTM \cite{hochreiter1997lstm} &
              2.17 &
              58.48 &
              \textbf{58.31} &
              60.35 &
              59.48 &
              59.74 &
              60.37 &
              59.52 &
              \textbf{59.39} \\
             &
              TFN \cite{zadeh2017_TFN} &
              LSTM \cite{hochreiter1997lstm} &
              5.04 &
              37.15 &
              20.12 &
              37.15 &
              20.12 &
              37.15 &
              20.12 &
              37.15 &
              20.12 \\
             &
              BERT\_MAG \cite{rahman2020_bertmag} &
              BERT \cite{kenton2019bert} &
              110.83 &
              62.83 &
              48.50 &
              61.39 &
              49.70 &
              62.83 &
              48.51 &
              62.35 &
              48.90 \\
             &
              LMF \cite{Liu2018_LMF} &
              LSTM \cite{hochreiter1997lstm} &
              1.10 &
              42.38 &
              34.48 &
              57.15 &
              57.85 &
              55.94 &
              56.63 &
              51.82 &
              49.65 \\
             &
              \textbf{Adapted (Ours)} &
              Transformer \cite{vaswani2017attention} &
              2.60 &
              \textbf{62.85} &
              55.55 &
              \textbf{62.49} &
              \textbf{60.00} &
              \textbf{63.32} &
              \textbf{60.69} &
              \textbf{62.89} &
              58.75\\
            \bottomrule
        \end{tabular}
    }
    \label{tab:mm-sentiment-other-methods}
\end{table}

\begin{table}[t]
    \centering
    \caption{Performance (top-1 accuracy) comparison with existing methods for action recognition on NTU RGB+D dataset. RGB and Depth columns report performance when only RGB and only Depth are available. Avg column indicates average performance. $^\ast$ indicates that available code and pretrained models were used to generate the results.}
    \label{tab:ntu-action-recognition}
    \setlength{\tabcolsep}{6pt}
    \centering
    \begin{tabular}{llccc}
        \toprule
        \multicolumn{1}{c}{Method} & \multicolumn{1}{c}{Backbone} & RGB            & Depth          & Avg        \\
        \midrule
        \midrule
        Modality Distill. \cite{garcia2018modalitydistill}      & ResNet-50 \cite{he2016resnet}      & 73.42          & 70.44          & 71.93          \\
        Luo et al. \cite{luo2018luoetal}      & ResNet-18 \cite{he2016resnet} + GRU \cite{cho-etal-2014GRU} & 89.50          & 87.50          & 88.50          \\
        DMCL \cite{garcia2021dmcl}            & ResNet-18 \cite{he2016resnet}             & 83.61          & 80.56          & 82.09          \\
        Motion-RGBD \cite{zhou2022motionrgbd} & DSN + DTN \cite{zhou2022motionrgbd}                    & 90.30          & 92.70          & 91.50          \\
        ActionMAE \cite{woo2023actionmae}     & ResNet-34 \cite{he2016resnet} + Transformer \cite{vaswani2017attention}    & 84.50          & 90.50          & 87.50          \\
        UMDR \cite{zhou2023umdr}$^\ast$       & DSN + DTN \cite{zhou2022motionrgbd}                   & 90.47          & 93.99          & 92.23          \\
        \textbf{Adapted (Ours)}    & DSN + DTN \cite{zhou2022motionrgbd}                   & \textbf{91.53} & \textbf{94.29} & \textbf{92.91} \\
        \bottomrule
    \end{tabular}
\end{table}

\subsection{Experiments on Multimodal Sentiment Analysis}
\label{sec:result-sentiment-analysis}
We tested our adaptation method for multimodal sentiment analysis on CMU-MOSI \cite{zadeh2016mosi} and CMU-MOSEI \cite{bagher2018mosei} datasets, and report the results in Table~\ref{tab:mm-sentiment-other-methods}. We use multimodal transformer (MulT) \cite{tsai2019mult} as the base model and adapt it using our approach.
We observed that when text is available and either audio or video or both are missing at the test time, the performance does not drop significantly. 
Similar trend was reported in \cite{hazarika2022robust_mmt2}. If text is missing at test time, then the performance of the base MulT model drops significantly. The Adapted models can partially compensate for missing modality and offer significantly better performance compared to the base MulT model. 

For CMU-MOSI dataset, we observe 1.69\% and 2.44\% improvement in accuracy and larger improvement in F1 score over the base MulT model when only audio and only visual are available, respectively. The adapted model offers significant improvement when audio-visual modalities are available and text is missing. It shows 6.56\% improvement in accuracy and 12.01\% improvement in F1 score over the base MulT model. For CMU-MOSEI dataset, we see even greater improvement in all the metrics. Experiments show 25.7\%, 24.21\% and 21.41\% improvement in accuracy for audio only, visual only and audio-visual scenarios compared to the MulT model. We also observe 27.91\%-36.30\% improvement in F1 score compared to the base MulT model.

We compare our adaptation method with existing methods for multimodal sentiment analysis. For CMU-MOSI dataset, BERT\_MAG works better in terms of accuracy but our adaptation method works better in terms of F1 score. One thing to mention is that BERT\_MAG uses a pretrained BERT model and finetunes it on the dataset but we are not using any pretraining on extra data. For CMU-MOSEI, our adaptation method works better for most of the cases.

\begin{table}[t]
    \centering
    \caption{Performance (accuracy) comparison with prompting based approach for multimodal classification on UPMC Food-101 dataset. Image and text columns indicate the amount of image and text modality available during both training and testing. $\dag$ indicates that those values are approximated from the plots published in \cite{lee2023map}.}
    \setlength{\tabcolsep}{6pt}
    \label{tab:food-101-sota-comparison}
    \begin{tabular}{cccccc}
        \toprule
        \multicolumn{2}{c}{Available Modality} & ViLT & Attention & Input & \textbf{Adapted}        \\
        Image & Text  &  \cite{kim2021vilt} & Prompts \cite{lee2023map} & Prompts \cite{lee2023map}       & \textbf{(Ours)}         \\
        \midrule
        \midrule
        100\% & 30\%  & 66.29~ & 72.57~   & 74.53~          & \textbf{75.38} \\
        30\%  & 100\% & 76.66~ & 86.05~   & 86.18~          & \textbf{88.31} \\
        65\%  & 65\%  & 69.25~ & 78.09~   & 79.08~          & \textbf{81.77} \\
        100\% & 0\%   & 63.60$^\dag$ & 67.70$^\dag$   & \textbf{68.10}$^\dag$ & 67.66          \\
        0\%   & 100\% & 76.10$^\dag$ & 85.30$^\dag$   & 84.80$^\dag$          & \textbf{86.01} \\
        \midrule
        \multicolumn{2}{l}{Average Accuracy} & 70.38~                 & 77.94~           & 78.54~       & \textbf{79.83} \\
        \midrule
        \multicolumn{2}{l}{Total Params (M)} & 112.26~                 & 112.49~           & 112.49~       & \textbf{112.47} \\
        \multicolumn{2}{l}{Learnable Params (M)} & 0.0~~                 & 0.221           & 0.221       & \textbf{0.207} \\
        \multicolumn{2}{l}{Change (\%)} & +0.0\%~~                 & +0.20\%~           & +0.20\%~       & \textbf{+0.18\%} \\
        \bottomrule
    \end{tabular}
\end{table}

\subsection{Experiments on Multimodal Action Recognition}
\label{sec:result-action-recognition}
We evaluate our approach on NTU RGB+D \cite{shahroudy2016ntudataset} dataset for multimodal action recognition task. We use UMDR \cite{zhou2023umdr} as the base model and adapt it using
our approach. As shown in Table~\ref{tab:ntu-action-recognition}, our adaptation performs better compared to recent modality masking and generation based approach ActionMAE \cite{woo2023actionmae} and modality de- and re-coupling based approaches Motion-RGBD \cite{zhou2022motionrgbd} and UMDR \cite{zhou2023umdr}. Our adaptation shows 7.03\% and 1.06\% improvement over ActionMAE and UMDR respectively when RGB is available and depth is missing. We see 3.79\% and 0.30\% improvement over ActionMAE and UMDR respectively when depth is available and RGB is missing. Moreover, our method outperforms all the existing baseline methods in all the scenarios. Which also indicates that our approach can learn better feature representation compared to modality masking, generation and distillation based approaches. 

The base UMDR model has 75.82M parameters. Our adaptation method adds 0.24M additional learnable parameters, which is only 0.32\% of the total model parameters. Other methods in this table do not report their total parameter counts, so we omit the total parameters column for this table.

\subsection{Experiments on Multimodal Classification}
\label{sec:result-classification}
To further evaluate the effectiveness of our approach, we compare it with recent prompt based approach missing-aware prompts \cite{lee2023map} on UPMC Food-101 \cite{wang2015food101} dataset. The results are summarized in Table~\ref{tab:food-101-sota-comparison}. For fair evaluation, we use the same experimental setup and evaluation script as \cite{lee2023map} to evaluate performance on different available and missing modality scenarios. Image and text columns indicate the amount of image and text modality available during both training and testing. Our adaptation method outperforms prompting based approach for most of the scenarios. On an average, our approach shows 1.29\% improvement over the best prompting method and 9.45\% improvement over the base ViLT model. These results corroborate the fact that adapting models by intermediate feature modulation helps the model learn optimal feature representation to perform better on different missing modality scenarios.

\textbf{Efficiency on Parameters.} We keep the pre-trained ViLT backbone frozen and compare the additional learnable parameters required for the learnable prompts \cite{lee2023map} and our method. We require less additional parameters while performing better than both input level and attention level prompts. Thus our adaptation method shows greater parameter efficiency and effectiveness compared to prompt based approach.

\begin{figure}[t]
    \centering
    \hfill
    \subfloat[Multimodal material segmentation on MCubeS dataset. Available: \textcolor{green}{RGB} - Missing: \textcolor{red}{AoLP, DoLP, NIR}\label{fig:mcubes-rgb-cos-sim}]{
        \includegraphics[width=0.45\textwidth]{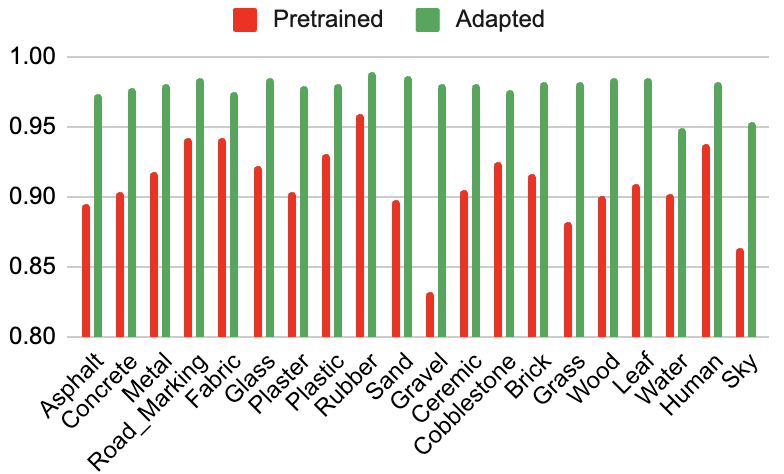}
    }
    \hfill
    \subfloat[Multimodal Action Recognition on NTU RGB+D Dataset. Available: \textcolor{green}{RGB} - Missing: \textcolor{red}{Depth}\label{fig:ntu-rgb-cos-sim-20}]{
        \includegraphics[width=0.45\textwidth]{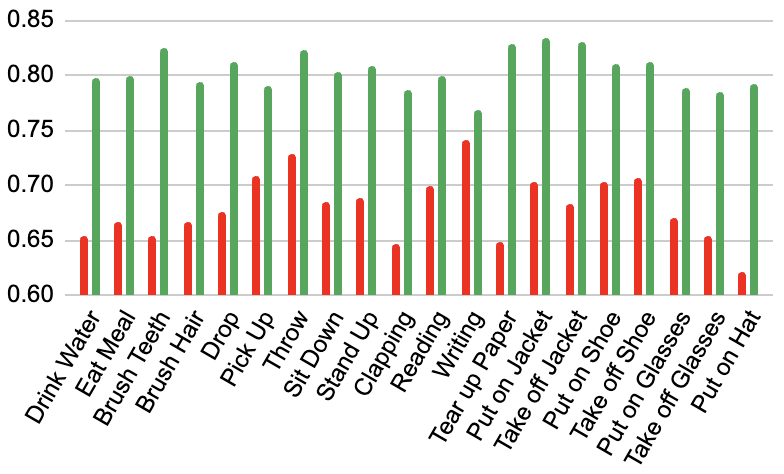}
    }
    \hfill
    \caption{Cosine similarity between complete and missing modality features of the pretrained model (Pretrained) and complete and missing modality features of the adapted model (Adapted) on MCubeS and NTU RGB+D datasets. Adapted models show higher similarity to the complete modality features compared to the pretrained model, indicating less deviation and better handling of missing modalities.}
    \label{fig:cos-sim}
\end{figure}

\subsection{Feature and Parameter Analysis}
We perform additional analysis to evaluate the effectiveness and generalizability of the adaptation approach. We discuss them in the section.

\subsubsection{Why adapted model performs better?}
\label{sec:mcubes-cosine-sim}
To further analyze how the adaption is helping the model improve overall performance, we conducted cosine similarity analysis of the final fused features extracted from the last layer of the network. To be specific, we calculate the cosine similarity between the complete and missing modality features from the pretrained model (Pretrained), and the cosine similarity between the complete and missing modality features from the adapted model (Adapted). We show the cosine similarities for each class in Figure~\ref{fig:cos-sim}.

The adapted model demonstrates a higher cosine similarity to the complete modality features compared to the pretrained model on both MCubeS dataset for multimodal material segmentation and NTU RGB+D dataset for multimodal action recognition when RGB is available and other modalities are missing. This increased similarity indicates that the adapted model better retains the essential information from the original complete modality input features, even when some modalities are missing. Consequently, this robustness in feature representation leads to a significant improvement in the model's overall performance. These results demonstrate the effectiveness of the adapted model in handling scenarios with missing modalities and maintaining robust prediction quality. 

We only show first 20 out of 60 classes for NTU RGB+D dataset here. We have included comparison for all the 60 classes including other missing scenarios in Section~\ref{sec:cos-sim-analysis-supp} in the supplementary materials.

\subsubsection{Performance gain vs learnable parameters}
\label{sec:performance-vs-params}
Our method achieves significant performance gains with a small number of  additional learnable parameters. As shown in Table~\ref{tab:MFNet-comparison-with-other-models} and \ref{tab:NYU-comparison-with-other-models}, adapted models provide 8.46\% and 16.54\% improvement in mIoU on an average over the base CMNeXt model with only 0.79M additional parameters (i.e., 0.68\% of the total model parameters). 
For multimodal sentiment analysis, as shown in Table~\ref{tab:mm-sentiment-other-methods}, adapted models provide 3.57\% and 23.78\% improvement in accuracy and 6.79\% and 33.22\% improvement in F1 score for CMU-MOSI and CMU-MOSEI datasets, respectively over the base MulT model with only 0.02M additional parameters (i.e., 0.775\% of the total model parameters). 
For multimodal classification on UPMC Food-101 dataset, as shown in Table~\ref{tab:food-101-sota-comparison}, adapted models achieve an average performance improvement of 9.45\% over the base ViLT model with only 0.207M additional learnable parameters (i.e., 0.18\% of the total model parameters). 

In summary, learning a small number of additional parameters in a base network provides significant performance improvement in the case of missing modalities across all tasks and architectures in our experiments. The parameter complexity of our approach is comparable/better than existing robust methods like CRM \cite{shin2023crm} in Table~\ref{tab:MFNet-comparison-with-other-models} and prompts \cite{lee2023map} in Table~\ref{tab:food-101-sota-comparison}. However, existing works on missing modality robustness vary widely in terms of model architectures \cite{fan2023spidermesh, wang2022tf}, fusion methods \cite{lin2023vpfnet, zhou2023umdr}, training procedures \cite{shin2023crm, fan2023spidermesh}, and missing feature generation methods \cite{woo2023actionmae}. Due to this heterogeneity, a fair comparison based solely on model size/number of parameters is infeasible. 

\section{Limitations and Future Directions}
In this work, our main focus was to enhance missing modality robustness of existing multimodal models. Though our method can make existing models robust to different missing modality scenarios, it has certain limitations. First, we only considered missing modality during test time. However in real life scenarios, modalities can be missing in both train and test times. 
Second, our method learns one set of adaptation parameters for every combination of missing modalities. While the number of adaptation parameters is small, the overall parameter complexity will scale with the number of modality combinations. For $M$ modalities, we can have up to $2^M$ possible combinations, as each modality can either be available or missing. Our method will require $2^M-2$ sets of adaptation parameters to accommodate every possible combination of missing modalities (excluding two cases when all or none of the modalities are available). If we expect one modality out of $M$ to be missing at the test time, which is the case in most of the published work, our method will require $M$ sets of adaptation parameters.
Third, we insert the learnable layers after each layer of the encoders and the fusion block. We did not try to optimize the number of parameters or find the optimal places to insert those learnable layers. Future study will explore these areas to further reduce the number of parameters, enhance the effectiveness and applicability of the approach in newer tasks and datasets.

\section{Conclusion}
Missing modalities at test time can cause significant degradation in the performance of multimodal systems. In this paper, we presented a simple and parameter-efficient adaptation method for robust multimodal learning with missing modalities. We demonstrated that simple linear operations can efficiently transform a single pretrained multimodal network and achieve performance comparable to multiple (independent) dedicated networks trained for different modality combinations. We evaluated the performance of our method and compared with existing robust methods for five different multimodal tasks. Our method requires an extremely small number of additional parameters (e.g., $<1\%$ of the total parameters in most experiments), while significantly improving performance compared to existing baseline models and methods for different missing modality scenarios. Our adaptation strategy is applicable to different network architectures, modalities and tasks, which can be a versatile solution to build robust multimodal systems.

\section*{Acknowledgements}
    This work is supported in part by AFOSR award FA9550-21-1-0330 and NSF CAREER award CCF-2046293. This work used Indiana Jetstream2 through allocation CIS220128 from the ACCESS program supported by NSF grants 2138259, 2138286, 2138307, 2137603, and 2138296.
 
\medskip
\bibliography{main}
\bibliographystyle{unsrt}

\clearpage
\newpage
\vfill

\newpage

\setcounter{page}{1}

\newcommand{\beginsupplement}{%
        \setcounter{table}{0}
        \renewcommand{\thetable}{S\arabic{table}}%
        \setcounter{figure}{0}
        \renewcommand{\thefigure}{S\arabic{figure}}%
        \setcounter{section}{0}
        \renewcommand{\thesection}{S\arabic{section}}%
     }

\beginsupplement
\begin{center}
    {\Large \sc Supplementary Material}
\end{center}

\setcounter{section}{0}
\section{Datasets}
\label{supp-datasets}

\hspace{\parindent}\textbf{MFNet Dataset} introduced by \cite{Ha2017mfnet}, is a popular dataset for RGB-thermal urban scene segmentation, particularly in the context of supporting autonomous driving applications. It comprises a total of 1569 aligned pairs of RGB-thermal images. Within this collection, 820 image pairs were captured during daytime, while 749 pairs were acquired during nighttime. The dataset is  divided into distinct training and test sets, each accompanied by pixel-level annotations that define semantic labels for nine classes. Each image is $640 \times 480$ pixels. %

\textbf{NYU Depth v2 (NYUDv2) Dataset} from \cite{Silberman2012nyudv2} is a well-known dataset for RGB-D semantic segmentation. This dataset contains 1449 pairs of aligned RGB-depth images of indoor scenes. The images are divided into training and test sets containing 795 and 654 pairs of images respectively. The dataset also provides per pixel annotations for 13 classes, 40 classes and 894 classes ground truth semantic labels. For our experiments we used the standard 40 classes annotation. Each image is $640 \times 480$ pixels and the dataset contains both raw and processed depth maps. For our experiments we used HHA images as proposed by \cite{gupta2014hha} instead of depth maps. %

\textbf{Multimodal Material Segmentation (MCubeS) Dataset} was introduced by \cite{Liang2022mcubes} for accurate multimodal material segmentation with the help of thermal and polarized images alongside RGB images. This dataset has four modalities: RGB, Angle of Linear Polarization, Degree of Linear Polarization and Near-Infrared. Alongside these modalities, the dataset also provides ground truth annotation for semantic and material segmentation. There are 500 image sets divided into train, validation and test sets having 302, 96 and 102 image sets respectively. The images are $1224 \times 1024$ pixels each and have 20 classes in total. 

\textbf{NTU RGB+D (NTU60)} dataset \cite{shahroudy2016ntudataset} is a popular multimodal action recognition dataset. The dataset contains 56,880 action samples divied into 60 classes. The actions can be braodly categorized into three different categories: daily actions, medical conditions and mutual actions. It has four different input modalities: RGB videos $(1920 \times 1080)$, depth map sequences $(512 \times 424)$, infrared (IR) videos $(512 \times 424)$ and 3D skeletal data (25 major body joints). Three Microsoft Kinect V2 cameras were used to capture the videos simultaneously. It has two evaluation protocols: cross subject and cross view. We used RGB and depth data for our experiments and evaluated on cross subject protocol.

\textbf{CMU-MOSI} dataset from \cite{zadeh2016mosi} is a popularly used for multimodal sentiment analysis. The dataset has 2199 samples each having audio, visual and text as input modalities. It is divided into train, validation and test sets containing 1284, 229 and 686 samples respectively along with annotated sentiment for each sample.

\textbf{CMU-MOSEI} is a large scale sentiment analysis dataset from \cite{bagher2018mosei}. It is 10 times larger than CMU-MOSI and contains audio, visual and text modalities along with ground truth sentiment annotations. The dataset contains 23453 samples divided into train, validation and test sets for multimodal sentiment analysis and emotion recognition.

\textbf{UPMC Food-101} dataset \cite{wang2015food101} is a popular challenging multimodal classification dataset. It has 90,704 image-text pairs divided into train, validation and test sets. The dataset is annotated for 101 classes. Classes are identical to the ETHZ Food-101 dataset \cite{bossard2014ETHZfood}. The samples are noisy as they were collected in an uncontrolled environment and thus huge diversity among samples is observed.

\section{Implementation Details}
\label{sec:impl-details-supplementary}
We used Python\footnote{\url{https://www.python.org/}} 3.8.12 and PyTorch\footnote{\url{https://pytorch.org/}} 1.9.0 to for our implementation. The experiments were done using two NVIDIA RTX 2080 Ti GPUs. We applied automatic mixed precision (AMP) training provided by PyTorch. For CMNeXt model, we use their publicly available code\footnote{\url{https://github.com/jamycheung/DELIVER}} and models trained on all the available modalities for each dataset. We trained the multimodal transformer models on all the modalities using the available code and preprocessed data from the repository\footnote{\label{mmsa}\url{https://github.com/thuiar/MMSA}} for CMU-MOSI and CMU-MOSEI datasets.

\textbf{MFNet Dataset:} We divided the 4 channel RGB-T images into three channel RGB and one channel thermal images. Then data pre-processing and augmentation was applied following CMNeXt from \cite{zhang2023cmnext}. MiT-B4 from \cite{xie2021segformer} was the backbone for the base CMNeXt model. One set of scale and shift parameters was learnt for each input modality combination. Input images were sized at $640 \times 480$ for both training and testing and we report single scale performance for all the experiments. The scale and shift parameters were trained for 100 epochs with a batch size of 4.

\begin{table}[t]
    \centering
    \caption{Hyperparameters for the experiments on CMU-MOSI and CMU-MOSEI datasets for multimodal sentiment analysis.}
    \setlength{\tabcolsep}{6pt}
    \begin{tabular}{lcc}
        \toprule
        \multicolumn{1}{c}{Hyperparameters} & CMU-MOSI & CMU-MOSEI \\
        \midrule
        \midrule
        Batch Size                          & 16       & 4         \\
        Initial Learning Rate               & 0.002    & 0.0005    \\
        Optimizer                           & Adam     & Adam      \\
        Attention Dropout                   & 0.3      & 0.4       \\
        Embedding Dropout                   & 0.2      & 0.0       \\
        Output Dropout                      & 0.5      & 0.5       \\
        Gradient Clip                       & 0.6      & 0.6       \\
        Weight Decay                        & 0.005    & 0.001     \\
        Temporal Conv Kernel Size (T/A/V)   & 5/5/5    & 5/1/3     \\
        \# of Crossmodal Blocks              & 4        & 4         \\
        \bottomrule
    \end{tabular}
    \label{tab:hyperparameters-mosi-mosei}
\end{table}

\begin{table}[th]
    \centering
    \caption{Learnable parameter counts for different parameter efficient model adaptation methods. As seen from the table, scale and shift introduce less than 0.7\% of the total model parameters.}
    \setlength{\tabcolsep}{6pt}
    \begin{tabular}{lccc}
        \toprule
        \multicolumn{1}{c}{\multirow{2}{*}{Method}} & Total & Learnable & \% of Total \\
        & Parameters (M) & Parameters & Parameters \\
        \midrule
        \midrule
        Norm            & 116.560 & 0.126 & 0.108 \\
        BitFit          & 116.560 & 0.378 & 0.324 \\
        LoRA            & 116.957 & 0.397 & 0.340 \\
        Scale and Shift & 117.349 & 0.789 & 0.673 \\
        \bottomrule
    \end{tabular}
    \label{tab:ssf-parameter-count}
\end{table}

\begin{table*}[t]
    \centering
    \caption{Performance comparison (\% mIoU) of different parameter-efficient adaptation techniques for MFNet, NYUDv2, and MCubeS datasets. Each column reports mIoU of the Adapted model with the corresponding modalities, and Avg indicates average performance. A and D denote Angle and Degree of Linear Polarization.}
    \setlength{\tabcolsep}{4pt}
    \begin{tabular}{l|ccc|ccc|cccc}
        \toprule
        \multicolumn{1}{c}{Datasets} & \multicolumn{3}{c}{MFNet}  & \multicolumn{3}{c}{NYUDv2}  & \multicolumn{4}{c}{MCubeS}                                        \\
        \midrule
        Methods  & RGB            & Thermal      & Avg & RGB            & Depth      & Avg & RGB            & RGB-A       & RGB-A-D  & Avg        \\ 
        \midrule
        \midrule
        Pretrained                   & 53.71 & 35.48 & 44.60 & 51.19 & 5.26 & 28.23 & 42.32          & 48.81          & 49.06          & 46.73          \\
        Dedicated           & \textbf{55.86} & \textbf{53.34} & \textbf{54.60} & 52.18 & 33.49 & 42.84 & 48.16          & 48.42          & 49.48          & 48.69          \\
        \midrule
        Scale Only                   & 54.77 & 49.23 & 52.00 & 53.04 & 36.12 & 44.58 & 50.16          & 50.55          & \textbf{51.13} & 50.61          \\
        Shift Only                   & 54.57 & 48.96 & 51.77 & 53.04 & 36.25 & 44.65 & 50.13          & 50.40          & 50.86          & 50.46          \\
        BitFit                       & 54.39 & 49.07 & 51.73 & \textbf{53.09} & 36.64 & \textbf{44.87} & 50.19          & 50.57          & 51.07          & 50.61          \\
        LoRA                         & 54.19 & 47.45 & 50.82 & 52.87 & 34.97 & 43.92 & 49.59          & 50.07          & 50.80          & 50.15          \\
        Norm                         & 54.65 & 47.49 & 51.07 & 53.05 & 34.73 & 43.49 & 49.95          & 50.51          & 51.07          & 50.51          \\
        \textbf{Scale and Shift}              & \textbf{55.22} & \textbf{50.89} & \textbf{53.06} & 52.82 & \textbf{36.72} & 44.77 & \textbf{50.43} & \textbf{50.62} & 51.11          & \textbf{50.72} \\ 
        \bottomrule
    \end{tabular}
    \label{tab:overall-comparison-with-other-methods}
\end{table*}

\begin{table}[t]
    \centering
    \caption{Performance comparison with parameter efficient model adaptation techniques on CMNeXt model for MFNet dataset. Average column indicates average performance when one of the two modalities gets missing. Mean accuracy, F1 score and \% mIoU are shown for all the experiments.}
    \setlength{\tabcolsep}{4pt}
    \centering
    \begin{tabular}{lccccccccc}
        \toprule
        \multicolumn{1}{c}{\multirow{2}{*}{Methods}} & \multicolumn{3}{c}{RGB} & \multicolumn{3}{c}{Thermal} & \multicolumn{3}{c}{Average} \\
        \multicolumn{1}{c}{}                         & mAcc   & F1    & \% mIoU   & mAcc     & F1    & \% mIoU     & mAcc      & F1     & \% mIoU   \\
        \midrule
        \midrule
        Pretrained                       & 60.74  & 66.91 & 53.71  & 38.18    & 45.11 & 35.48    & 49.46    & 56.01   & 44.60 \\
        Dedicated                           & 66.28  & \textbf{68.22} & \textbf{55.86}  & \textbf{68.35}    & \textbf{65.29} & \textbf{53.34}    & \textbf{67.32}    & \textbf{66.76}   & \textbf{54.60} \\
        \midrule
        Scale Only                                   & 67.09  & 68.03 & 54.77  & 64.00    & 60.92 & 49.23    & 65.55    & 64.48   & 52.00 \\
        Shift Only                                   & 65.82  & 67.42 & 54.57  & 59.77    & 60.54 & 48.96    & 62.80    & 63.98   & 51.77 \\
        BitFit                                       & 66.49  & 67.40 & 54.39  & 61.06    & 60.59 & 49.07    & 63.78    & 64.00   & 51.73 \\
        LoRA                                         & 66.44  & 67.32 & 54.19  & 57.10    & 59.04 & 47.45    & 61.77    & 63.18   & 50.82 \\
        Norm                                         & 66.43  & 67.07 & 54.65  & 57.55    & 59.22 & 47.49    & 61.99    & 63.15   & 51.07 \\
        \textbf{Scale and Shift}                              & \textbf{67.18} & 68.04 & \textbf{55.22} & \textbf{66.70} & \textbf{62.64} & \textbf{50.89} & \textbf{66.94} & \textbf{65.34} & \textbf{53.06} \\
        \bottomrule
    \end{tabular}
    \label{tab:MFNet-comparison-with-other-methods}
\end{table}

\textbf{NYUDv2 Dataset:} For processing depth maps, we follow SA-Gate by \cite{chen2020SAGate} and CMNeXt by \cite{zhang2023cmnext} and use HHA-encoded images instead of raw depth maps. The already preprocessed dataset can be downloaded from the SA-Gate repository\footnote{\url{https://github.com/charlesCXK/RGBD_Semantic_Segmentation_PyTorch}}. RGB and HHA images were sized at $640 \times 480$ pixels each and we used this size for training and testing. The backbone was set to MiT-B4 as suggested in CMNeXt paper. One set of scale and shift parameters was learnt for each input modality combination by feeding available input modalities and setting the missing modality to zero. We train the scale and shift parameters for 100 epochs with a batch size of 4 and report single scale performance.

\textbf{MCubeS Dataset:} We follow the same data pre-processing and augmentations used by the base CMNeXt model from \cite{zhang2023cmnext}. MiT-B2 from \cite{xie2021segformer} was used as the backbone for this dataset. We set the input image resolution to $512 \times 512$ during training and $1024 \times 1024$ during testing and report single scale performance with predicted segmentation maps sized at $1024 \times 1024$. Similar to other two datasets, we train the learnable parameters for 100 epochs with a batch size of 4. 

\textbf{NTU RGB+D (NTU60) Dataset:} We followed the same pre-processing and experimental setup as \cite{zhou2022motionrgbd, zhou2023umdr}. We used RGB and depth data for our experiments and evaluated our method using cross subject protocol for fair comparison with \cite{luo2018luoetal, woo2023actionmae, zhou2022motionrgbd, zhou2023umdr}. We extract 16 frames per video following previous methods and utilize DSN and DTN for spatial and temporal information encoding following \cite{zhou2022motionrgbd, zhou2023umdr}. Then we train the SSF layers for 20 epochs with a batch size of 6.

\textbf{CMU-MOSI and CMU-MOSEI Datasets:} We used Multimodal Transformer (MulT) from \cite{tsai2019mult} as the base model. Preprocessed datasets and all the configurations are available on the repository\footnote{\url{https://github.com/thuiar/MMSA}}. First we trained the multimodal transformer (MulT) model on all the available modalities and then adapted the pretrained model for different missing modality scenarios. The hyperparameters for the experiments are shown in Table~\ref{tab:hyperparameters-mosi-mosei}.

\textbf{UPMC Food-101 Dataset:} For multimodal classification on this dataset, we use ViLT as the base model and follow the experimental setup used by \cite{lee2023map}. We use the same hyper-parameters and script for generating and evaluating different missing modality combinations. The SSF layers were trained for 10 epochs with a learning rate of $1e^{-5}$.

\section{Number of Learnable Parameters}
We report the number of learnable parameters for different parameter-efficient adaptation techniques (for multimodal segmentation) in Table~\ref{tab:ssf-parameter-count}. We insert scale and shift layers after each linear, convolutional and norm (both batch norm and layer norm) layers. The number of learnable parameter varies with the size of the backbone. We used MiT-B4 as the backbone while counting these learnable parameters. Scale and shift adds only 0.789M learnable parameters which is less than 0.7\% of the total model parameters. Despite this very few parameters, it improves performance significantly in different missing modality scenarios. For this study we mainly focused on improving missing modality robustness and did not try to optimize the number of learnable parameters. We will leave that part for future studies.

\section{Performance Comparison with Parameter Efficient Model Adaption Techniques}
\label{sec:other-peft-comparison}
We performed a detailed performance comparison with other parameter efficient model adaptation methods for the three segmentation datasets. Comparison among different parameter-efficient adaptation methods show that SSF-based adaptation provides overall best performance. We summarize the results for scale only, shift only, BitFit \mbox{\cite{zaken2022bitfit}}, norm layer fine-tuning and LoRA \mbox{\cite{hu2021lora}} in Table~\ref{tab:overall-comparison-with-other-methods}. We also show detailed comparison in Table~\ref{tab:MFNet-comparison-with-other-methods} for RGB-thermal segmentation on MFNet dataset, Table~\ref{tab:nyu-other-methods} for RGB-depth segmentation on NYUDv2 dataset and Table~\ref{tab:mcubes-other-methods} for multimodal material segmentation on MCubeS dataset. For each method, we take a model trained on all the available modalities. Then we freeze the pretrained weights and tune the learnable parameters for the corresponding adaption method. We have shown mean accuracy, F1 score and \% mIoU for each experiment.

\begin{table}[t]
    \centering
    \caption{Performance comparison with parameter efficient model adaptation techniques on CMNeXt model for NYUDv2 dataset. Average column indicates average performance when one of the two modalities gets missing. Mean accuracy, F1 score and \% mIoU are shown for all the experiments.}
    \setlength{\tabcolsep}{6pt}
    \centering
    \begin{tabular}{lccccccccc}
        \toprule
        \multicolumn{1}{c}{\multirow{2}{*}{Methods}} & \multicolumn{3}{c}{RGB} & \multicolumn{3}{c}{Depth} & \multicolumn{3}{c}{Average} \\
        \multicolumn{1}{c}{}                         & mAcc   & F1    & \% mIoU   & mAcc   & F1   & \% mIoU      & mAcc   & F1    & \% mIoU       \\
        \midrule
        \midrule
        Pretrained                       & 64.10  & 65.70 & 51.19  & 8.30   & 7.95  & 5.26     & 36.20  & 36.83 & 28.23      \\
        Dedicated                           & 66.00  & 66.62 & 52.18  & 44.80  & 46.79 & 33.49    & 55.40  & 56.71 & 42.84      \\
        \midrule
        Scale Only                                   & \textbf{68.18}  & \textbf{67.38} & 53.04  & 51.54  & 49.88 & 36.12    & 59.86  & 58.63 & 44.58      \\
        Shift Only                                   & 67.54  & 67.35 & 53.04  & 50.30  & 49.76 & 36.25    & 58.92  & 58.56 & 44.65      \\
        BitFit                                       & 67.31  & 67.33 & \textbf{53.09}  & 50.68  & 50.27 & 36.64    & 59.00  & 58.80 & \textbf{44.87}      \\
        LoRA                                         & 66.67  & 67.14 & 52.87  & 49.34  & 48.66 & 34.97    & 58.01  & 57.90 & 43.92      \\
        Norm                                         & 67.18  & 67.34 & 53.05  & 48.74  & 48.06 & 34.73    & 57.96  & 57.70 & 43.89     \\
        \textbf{Scale and Shift}                                         & 67.96  & 67.18 & 52.82 & \textbf{52.42} & \textbf{50.60} & \textbf{36.72} & \textbf{60.19} & \textbf{58.89} & 44.77 \\
        \bottomrule
    \end{tabular}
    \label{tab:nyu-other-methods}
\end{table}

\begin{table}[t]
    \setlength{\tabcolsep}{2.5pt}
    \centering
    \caption{Performance comparison with different parameter efficient model adaptation techniques on CMNeXt model for MCubeS dataset. Average column indicates the average performance. Mean accuracy, F1 score and \% mIoU are shown for all the experiments.}
        \begin{tabular}{lcccccccccccc}
            \toprule
            \multicolumn{1}{c}{\multirow{2}{*}{Methods}} & \multicolumn{3}{c}{RGB} & \multicolumn{3}{c}{RGB-AoLP} & \multicolumn{3}{c}{RGB-AoLP-DoLP} & \multicolumn{3}{c}{Average} \\
            \multicolumn{1}{c}{}   & mAcc   & F1    & \% mIoU  & mAcc   & F1    & \% mIoU  & mAcc  & F1             & \% mIoU     & mAcc   & F1    & \% mIoU   \\
            \midrule
            \midrule
            Pretrained & 51.63 & 55.91 & 42.32 & 58.66 & 62.00 & 48.81 & 60.06   & 62.43          & 49.06    & 56.78  & 60.11 & 46.73  \\
            Dedicated     & 57.70 & 60.95 & 48.16 & 57.56 & 61.17 & 48.42 & 59.12   & 61.91          & 49.48    & 58.13  & 61.34 & 48.69  \\
            \midrule
            Scale Only             & 59.64 & 63.06 & 50.16 & 60.28 & 63.55 & 50.55 & 60.96   & \textbf{64.14} & \textbf{51.13}    & 60.29 & 63.58 & 50.61  \\
            Shift Only             & 59.82 & 63.17 & 50.13 & 60.10 & 63.36 & 50.40 & 60.61   & 63.78          & 50.86    & 60.18  & 63.44 & 50.46  \\
            BitFit                 & 59.98 & 63.24 & 50.19 & 60.12 & 63.52 & 50.57 & 60.84   & 64.03          & 51.07    & 60.31  & 63.60 & 50.61  \\
            LoRA                   & 59.08 & 62.50 & 49.59 & 59.81 & 63.05 & 50.07 & 60.69   & 63.84          & 50.80    & 59.86  & 63.13 & 50.15  \\
            Norm                   & 59.57 & 62.89 & 49.95 & 60.22 & 63.49 & 50.51 & \textbf{60.98} & 64.08   & 51.07    & 60.26  & 63.49 & 50.51  \\
            \textbf{Scale and Shift} & \textbf{60.23} & \textbf{63.41} & \textbf{50.43} & \textbf{60.40} & \textbf{63.59} & \textbf{50.62} & 60.94 & 64.04 & 51.11 & \textbf{60.52} & \textbf{63.68} & \textbf{50.72} \\
            \bottomrule
        \end{tabular}
    \label{tab:mcubes-other-methods}
\end{table}

\begin{table*}[t]
    \centering
    \caption{Per class \% IoU comparison between pretrained and adapted CMNeXt model on MFNet dataset. Adapted model show better performance for most of the classes leading to overall performance improvement.}

    \setlength{\tabcolsep}{4pt}
    \centering
    \begin{tabular}{ll|ccccccccc|c}
        \toprule
          Modalities &
          Methods &
          \rotatebox[origin=c]{90}{Unlabeled} &
          \rotatebox[origin=c]{90}{Car} &
          \rotatebox[origin=c]{90}{Person} &
          \rotatebox[origin=c]{90}{Bike} &
          \rotatebox[origin=c]{90}{Curve} &
          \rotatebox[origin=c]{90}{Car\_Stop} &
          \rotatebox[origin=c]{90}{Guardrail} &
          \rotatebox[origin=c]{90}{Color\_Cone} &
          \rotatebox[origin=c]{90}{Bump} &
          \rotatebox[origin=c]{90}{Mean} \\
        \midrule
        \midrule
        RGB-Thermal              & CMNeXt & 98.31 & 90.27 & 74.52 & 64.52          & 46.64 & 39.19 & 15.09          & 52.56 & 59.79 & 60.10 \\
        \midrule
        \multirow{2}{*}{RGB}     & Pretrained & \textbf{97.79} & 87.62 & 51.13 & \textbf{61.94} & 30.05 & 39.36 & \textbf{21.04} & 45.55 & 48.95 & 53.71 \\
         &
          \textbf{Adapted} &
          \textbf{97.79} &
          \textbf{88.06} &
          \textbf{55.55} &
          61.20 &
          \textbf{34.19} &
          \textbf{40.52} &
          15.78 &
          \textbf{48.67} &
          \textbf{55.21} &
          \textbf{55.22} \\
        \midrule
        \multirow{2}{*}{Thermal} & Pretrained & 95.97 & 55.24 & 68.47 & 9.27           & 31.85 & 2.75  & 0.0            & 16.87 & 38.92 & 35.48 \\
         &
          \textbf{Adapted} &
          \textbf{97.46} &
          \textbf{82.83} &
          \textbf{70.12} &
          \textbf{49.03} &
          \textbf{40.89} &
          \textbf{26.79} &
          \textbf{1.84} &
          \textbf{36.24} &
          \textbf{52.83} &
          \textbf{50.89} \\
        \midrule
    \end{tabular}
    \label{tab:per-class-iou-mfnet-cmnext-vs-ours}
\end{table*}

\begin{table}[t]
    \centering
    \caption{Per class \% IoU comparison between pretrained and adapted CMNeXt model on MCubeS dataset. Adapted model show better performance for most of the classes leading to overall performance improvement. Here A, D and N stand for Angle of Linear Polarization (AoLP), Degree of Linear Polarization (DoLP) and Near-Infrared (NIR) respectively.}
    \setlength{\tabcolsep}{2pt}
    \resizebox{\textwidth}{!}{
        \begin{tabular}{ll|cccccccccccccccccccc|c}
            \toprule
              Modalities &
              Methods &
              \rotatebox[origin=c]{90}{Asphalt} &
              \rotatebox[origin=c]{90}{Concrete} &
              \rotatebox[origin=c]{90}{Metal} &
              \rotatebox[origin=c]{90}{Road\_Marking} &
              \rotatebox[origin=c]{90}{Fabric} &
              \rotatebox[origin=c]{90}{Glass} &
              \rotatebox[origin=c]{90}{Plaster} &
              \rotatebox[origin=c]{90}{Plastic} &
              \rotatebox[origin=c]{90}{Rubber} &
              \rotatebox[origin=c]{90}{Sand} &
              \rotatebox[origin=c]{90}{Gravel} &
              \rotatebox[origin=c]{90}{Ceramic} &
              \rotatebox[origin=c]{90}{Cobblestone} &
              \rotatebox[origin=c]{90}{Brick} &
              \rotatebox[origin=c]{90}{Grass} &
              \rotatebox[origin=c]{90}{Wood} &
              \rotatebox[origin=c]{90}{Leaf} &
              \rotatebox[origin=c]{90}{Water} &
              \rotatebox[origin=c]{90}{Human} &
              \rotatebox[origin=c]{90}{Sky} &
              \rotatebox[origin=c]{90}{Mean} \\
            \midrule
            \midrule
            RGB-A-D-N &
              CMNeXt &
              84.4 &
              44.9 &
              53.9 &
              74.6 &
              32.1 &
              54.0 &
              0.8 &
              28.7 &
              29.8 &
              67.0 &
              66.2 &
              27.7 &
              68.5 &
              42.8 &
              58.7 &
              49.7 &
              75.3 &
              55.6 &
              19.1 &
              96.52 &
              51.5 \\
            \midrule
            \multirow{2}{*}{RGB} &
              Pretrained &
              69.7 &
              39.2 &
              47.6 &
              67.3 &
              26.9 &
              44.6 &
              0.2 &
              20.9 &
              15.2 &
              61.8 &
              36.7 &
              19.1 &
              67.2 &
              36.0 &
              49.5 &
              36.1 &
              71.6 &
              36.1 &
              14.7 &
              86.3 &
              42.3 \\
             &
              \textbf{Adapted} &
              \textbf{85.8} &
              \textbf{43.7} &
              \textbf{52.6} &
              \textbf{73.8} &
              \textbf{27.9} &
              \textbf{51.0} &
              \textbf{0.8} &
              \textbf{24.2} &
              \textbf{30.4} &
              \textbf{67.8} &
              \textbf{72.9} &
              \textbf{27.1} &
              \textbf{68.1} &
              \textbf{42.9} &
              \textbf{57.6} &
              \textbf{49.0} &
              \textbf{74.9} &
              \textbf{43.4} &
              \textbf{18.3} &
              \textbf{96.5} &
              \textbf{50.4} \\
            \midrule
            \multirow{2}{*}{RGB-A} &
              Pretrained &
              83.2 &
              43.3 &
              50.7 &
              72.6 &
              26.4 &
              51.9 &
              0.2 &
              \textbf{28.1} &
              22.2 &
              67.7 &
              63.4 &
              22.7 &
              \textbf{67.5} &
              40.6 &
              54.4 &
              44.9 &
              73.9 &
              44.8 &
              \textbf{21.8} &
              96.0 &
              48.8 \\
             &
              \textbf{Adapted} &
              \textbf{84.4} &
              \textbf{45.4} &
              \textbf{53.8} &
              \textbf{74.5} &
              \textbf{30.4} &
              \textbf{53.2} &
              \textbf{0.6} &
              26.9 &
              \textbf{28.8} &
              \textbf{69.0} &
              \textbf{69.3} &
              \textbf{24.8} &
              \textbf{67.5} &
              \textbf{43.2} &
              \textbf{58.4} &
              \textbf{48.2} &
              \textbf{75.1} &
              \textbf{48.1} &
              14.4 &
              \textbf{96.4} &
              \textbf{50.6} \\
            \midrule
            \multirow{2}{*}{RGB-A-D} &
              Pretrained &
              \textbf{84.5} &
              41.2 &
              46.7 &
              72.8 &
              25.2 &
              51.6 &
              0.3 &
              26.1 &
              28.8 &
              66.7 &
              65.6 &
              \textbf{26.0} &
              66.5 &
              40.4 &
              50.0 &
              45.1 &
              72.7 &
              49.4 &
              \textbf{25.6} &
              96.3 &
              49.1 \\
             &
              \textbf{Adapted} &
              84.1 &
              \textbf{45.6} &
              \textbf{54.1} &
              \textbf{74.6} &
              \textbf{30.5} &
              \textbf{54.2} &
              \textbf{0.6} &
              \textbf{28.1} &
              \textbf{30.1} &
              \textbf{69.0} &
              \textbf{67.6} &
              25.9 &
              \textbf{67.8} &
              \textbf{43.8} &
              \textbf{58.0} &
              \textbf{49.1} &
              \textbf{75.0} &
              \textbf{53.7} &
              13.7 &
              \textbf{96.5} &
              \textbf{51.1} \\
            \bottomrule
        \end{tabular}
    }
    \label{tab:per-class-iou-mcubes-cmnext-vs-ours}
\end{table}

\begin{figure}[t]
    \centering
    \hfill
    \subfloat[Available: \textcolor{green}{RGB, AoLP} - \textcolor{red}{Missing: DoLP, NIR}\label{fig:mcubes-rgba-cos-sim}]{
        \includegraphics[width=0.45\textwidth]{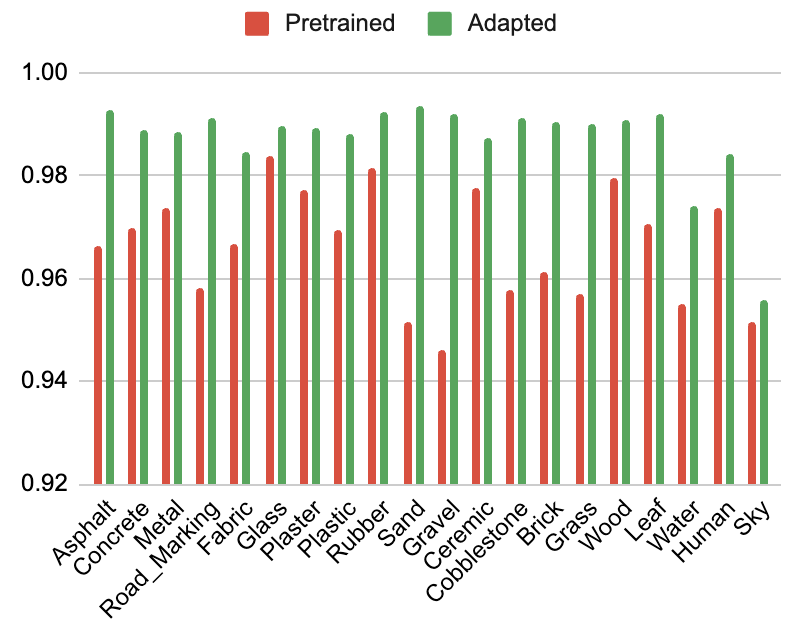}
    }
    \hfill
    \subfloat[Available: \textcolor{green}{RGB, AoLP, DoLP} - Missing: \textcolor{red}{NIR}\label{fig:mcubes-rgbad-cos-sim}]{
        \includegraphics[width=0.45\textwidth]{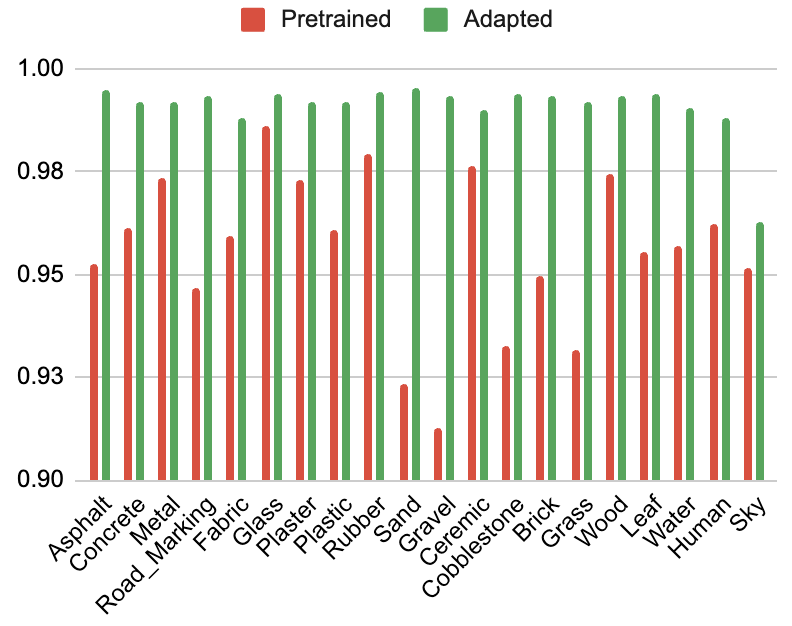}
    }
    \hfill
    \caption{Cosine similarity between complete and missing modality features of the pretrained model (Pretrained) and complete and missing modality features of the adapted model (Adapted) under different missing modality scenarios on MCubeS dataset. The adapted model shows higher similarity to the complete modality features compared to the pretrained model, indicating less deviation and better handling of missing modalities.}
    \label{fig:mcubes-cos-sim}
\end{figure}

\begin{figure}[t]
    \centering
    \subfloat[Available: \textcolor{green}{RGB} - Missing: \textcolor{red}{Depth}\label{fig:ntu-rgb-cos-sim}]{
        \includegraphics[width=0.90\textwidth]{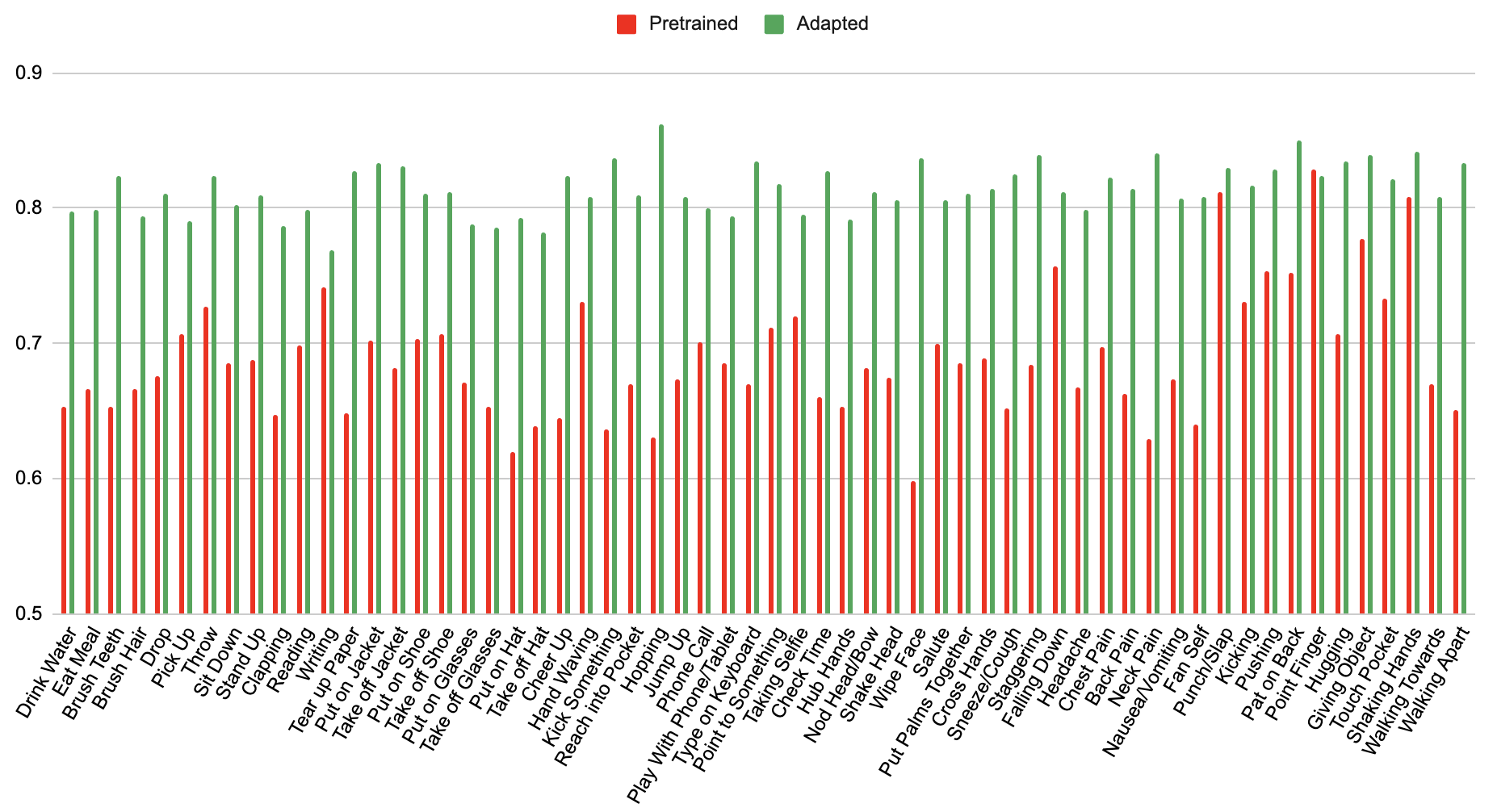}
    }
    
    \subfloat[Available: \textcolor{green}{Depth} - Missing: \textcolor{red}{RGB}\label{fig:ntu-depth-cos-sim}]{
        \includegraphics[width=0.90\textwidth]{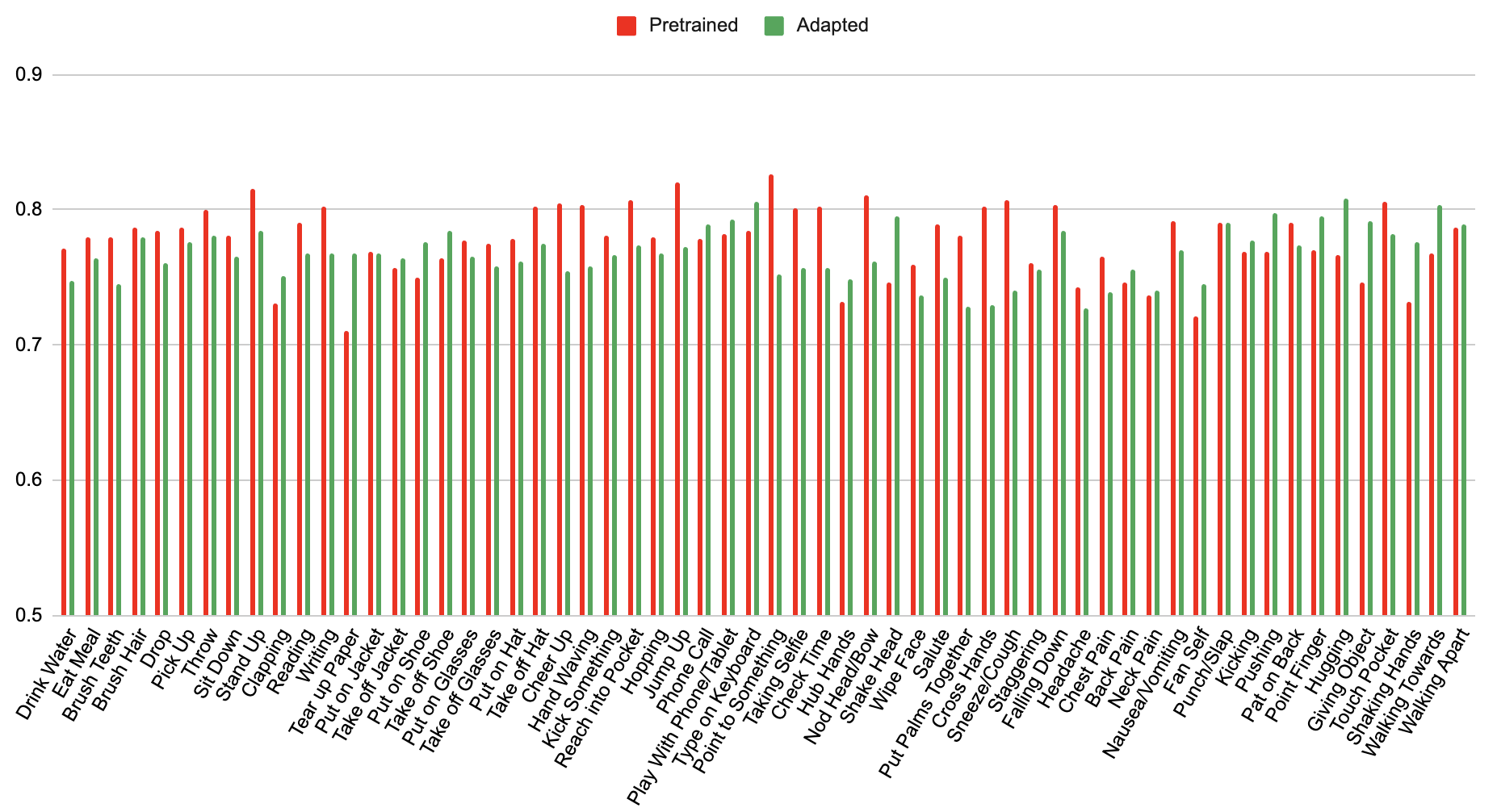}
    }
    \caption{Cosine similarity between complete and missing modality features of the pretrained model (Pretrained) and complete and missing modality features of the adapted model (Adapted) under different missing modality scenarios on NTU RGB+D dataset. Adapted models show comparable/higher similarity to the complete modality features compared to the pretrained model, indicating less deviation and better handling of missing modalities.}
    \label{fig:ntu-cos-sim}
\end{figure}

\begin{table}[t]
    \centering
    \caption{Missing modality performance comparison of base CMNeXt and Adapted CMNeXt model with Resnet-34 \cite{he2016resnet}, Resnet-50 \cite{he2016resnet}, and Swin-S \cite{liu2021swin} backbones on MFNet dataset.}
    \setlength{\tabcolsep}{4pt}
    \resizebox{\textwidth}{!}{
    \begin{tabular}{llccccccccc}
        \toprule
        \multicolumn{1}{c}{\multirow{2}{*}{\textbf{Method}}} &
          \multicolumn{1}{c}{\multirow{2}{*}{\textbf{Backbone}}} &
          \multirow{1}{*}{\textbf{Params}} &
          \multicolumn{2}{c}{\textbf{RGB-Thermal}} &
          \multicolumn{2}{c}{\textbf{RGB}} &
          \multicolumn{2}{c}{\textbf{Thermal}} &
          \multicolumn{2}{c}{\textbf{Missing Avg.}} \\
        \multicolumn{1}{c}{} &
          \multicolumn{1}{c}{} &
          \textbf{(M)}&
          \textbf{mAcc} &
          \textbf{\%mIoU} &
          \textbf{mAcc} &
          \textbf{\%mIoU} &
          \textbf{mAcc} &
          \textbf{\%mIoU} &
          \textbf{mAcc} &
          \textbf{\%mIoU} \\
        \midrule
        \midrule
        CMNeXt &
          ResNet-34  &
          50.72 &
          50.56 &	
          45.61 &
          23.95 &
          16.98 &
          19.25 &
          16.94 &
          21.60 &
          16.96 \\
        \textbf{Adapted CMNeXt } &
          ResNet-34 &
          50.91 &
          50.56 &	
          45.61 &
          \textbf{31.68} &
          \textbf{27.13} &
          \textbf{32.04} &
          \textbf{27.81} &
          \textbf{31.86} &
          \textbf{27.47} \\
        \midrule
        CMNeXt &
          ResNet-50&
          54.68 &
          50.69 &	
          46.05 &
          12.00 &
          11.18 &
          19.45 &
          18.38 &
          15.73 &
          14.78 \\
        \textbf{Adapted CMNeXt } &
          ResNet-50  &
          54.90 &
          50.69 &	
          46.05 &
          \textbf{39.84} &
          \textbf{34.69} &
          \textbf{28.70} &
          \textbf{25.59} &
          \textbf{34.27} &
          \textbf{30.14} \\
        \midrule
        CMNeXt &
          Swin-S &
          123.73 &
          45.24 &	
          41.02 &
          15.95 &
          11.98 &
          14.77 &
          13.91 &
          15.36 &
          12.95 \\
        \textbf{Adapted CMNeXt } &
          Swin-S &
          124.14 &
          45.24 &	
          41.02 &
          \textbf{39.69} &
          \textbf{34.11} &
          \textbf{28.01} &
          \textbf{25.05} &
          \textbf{33.85} &
          \textbf{29.58} \\
        \bottomrule
    \end{tabular}
    }
    \label{tab:cmnext-different-backbones}
\end{table}

\subsection{Performance comparison with other parameter-efficient model adaption techniques}
\label{sec:parameter-efficient-method-comparison}
Apart from robust models, we also compare different parameter-efficient adaptation techniques. We summarize the results in Table~\ref{tab:overall-comparison-with-other-methods}. For MFNet dataset, SSF outperforms all the methods and performance is significantly better than the Pretrained model and close to the Dedicated models. For NYUDv2 and MCubeS datasets, the Adapted model performs better than both Pretrained and Dedicated models. These experiments also show that SSF performs better than other methods for most of the input modality combinations for all the datasets. We show a detailed comparison for each dataset in terms of mean accuracy, F1 score and \% mIoU in Table~\ref{tab:MFNet-comparison-with-other-methods} - \ref{tab:mcubes-other-methods}.

\subsection{Performance Comparison for RGB-Thermal Semantic Segmentation on MFNet Dataset}
Table~\ref{tab:MFNet-comparison-with-other-methods} summarizes the results on MFNet dataset when the base CMNeXt model is adapted with other parameter efficient model adaptation techniques. Experiments show that scale and shift shows the best performance in all three matrices compared to all other methods. It shows a significant improvement of +8.46\% in mIoU, +9.33\% in F1 score and +17.48\% in mean accuracy on an average over the pretrained model. The average performance is also close to dedicatedly trained models. 

\subsection{Performance Comparison for RGB-Depth Semantic Segmentation on NYUDv2 Dataset}
Similar trend is observed for RGB-Depth semantic segmentation on NYUDv2 dataset as shown in Table~\ref{tab:nyu-other-methods}. Scale only and BitFit adapted models show slightly better performance for some of the matrices. But in most of the cases scale and shift adapted model performs better. For all the matrices, scale and shift shows a significant improvement of +16.54\% in mIoU, +22.05\% in F1 score and +23.99\% in mean accuracy over the pretrained model on an average and consistently outperforms dedicated training.

\subsection{Performance Comparison for Multimodal Material Segmentation on MCubeS Dataset}
We show comparison with parameter efficient model adaptation techniques on MCubeS dataset in Table~\ref{tab:mcubes-other-methods}. Scale and shift outperforms all other methods in most of the matrices for all input combinations. It also shows an improvement of +3.99\% in mIoU, +3.57\% in F1 score and +3.74\% in mean accuracy on an average over the pretrained model. Furthermore, Scale and shift also outperforms dedicated training for all input modality combinations. These experiments corroborate the fact that scale and shift provides better model adaption for different missing modality scenarios.

\section{Per Class IoU Comparison}
\label{sec:per-class-iou-supp}
To further analyze how the adaption is helping the model improve overall semantic and material segmentation performance, we conduct a per-class \% intersection over union (IoU) analysis on the pretrained and adapted models. Table~\ref{tab:per-class-iou-mfnet-cmnext-vs-ours} and \ref{tab:per-class-iou-mcubes-cmnext-vs-ours} summarize the results. We show the per class \% IoU comparison for different missing modality situations on MFNet dataset on Table~\ref{tab:per-class-iou-mfnet-cmnext-vs-ours}. From the table we can see that when RGB is available and thermal is missing, the adaptation helps improve performance for most of the classes. Though we see some performance drop for bike (-0.74\%) and guardrail (-5.25\%) classes, the rest of the classes have better \% IoU than the pretrained model. Bump (+6.26\%), person (+4.42\%), and curve (+4.14\%) classes show greater improvement after adaptation. When thermal is available and RGB is missing, adaptation improves performance for all the classes. Among the classes, bike (+39.76\%), car (+27.59\%), car stop (+24.04\%), color cone (+19.37\%) and bump (+13.91\%) are showing impressive performance improvement over the pretrained model.

\begin{table}[t]
    \centering
    \caption{Per-class IoU comparison among Dedicated, Pretrained, and Adapted model on MFNet Dataset.}
    \label{tab:mfnet-failure-case}
    \centering
    \begin{tabular}{lcccccc}
        \toprule
        \multicolumn{1}{c}{} &
          \multicolumn{3}{c}{RGB} &
          \multicolumn{3}{c}{Thermal} \\
        \multicolumn{1}{c}{\multirow{-2}{*}{Class}} &
          Dedicated &
          Pretrained &
          \textbf{Adapted} &
          Dedicated &
          Pretrained &
          \textbf{Adapted} \\
        \midrule
        \midrule
        Unlabeled &
          97.95 &
          97.79 &
          97.79 &
          97.85 &
          95.97 &
          97.45 \\
        Car &
          88.21 &
          87.62 &
          88.06 &
          85.84 &
          55.24 &
          83.39 \\
        Person &
          \cellcolor[HTML]{C0C0C0}62.97 &
          \cellcolor[HTML]{C0C0C0}{\color[HTML]{FE0000} 51.13} &
          \cellcolor[HTML]{C0C0C0}55.55 &
          71.00 &
          68.47 &
          69.96 \\
        Bike &
          63.03 &
          61.94 &
          61.20 &
          \cellcolor[HTML]{C0C0C0}56.50 &
          \cellcolor[HTML]{C0C0C0}{\color[HTML]{FE0000} 9.27} &
          \cellcolor[HTML]{C0C0C0}49.19 \\
        Curve &
          \cellcolor[HTML]{C0C0C0}36.02 &
          \cellcolor[HTML]{C0C0C0}{\color[HTML]{FE0000} 30.05} &
          \cellcolor[HTML]{C0C0C0}34.19 &
          40.03 &
          31.85 &
          40.53 \\
        Car\_Stop &
          40.45 &
          39.36 &
          40.52 &
          25.70 &
          2.75 &
          25.14 \\
        Guardrail &
          11.35 &
          21.04 &
          15.78 &
          \cellcolor[HTML]{C0C0C0}7.52 &
          \cellcolor[HTML]{C0C0C0}{\color[HTML]{FE0000} 0.00} &
          \cellcolor[HTML]{C0C0C0}1.55 \\
        Color\_Cone &
          \cellcolor[HTML]{C0C0C0}52.99 &
          \cellcolor[HTML]{C0C0C0}{\color[HTML]{FE0000} 45.55} &
          \cellcolor[HTML]{C0C0C0}48.67 &
          \cellcolor[HTML]{C0C0C0}42.24 &
          \cellcolor[HTML]{C0C0C0}{\color[HTML]{FE0000} 16.87} &
          \cellcolor[HTML]{C0C0C0}33.85 \\
        Bump &
          49.78 &
          48.95 &
          55.21 &
          53.35 &
          38.92 &
          53.22 \\
        \midrule
        Average &
          \textbf{55.86} &
          53.71 &
          55.22 &
          \textbf{53.34} &
          35.48 &
          50.48 \\
        \bottomrule
    \end{tabular}
\end{table}

\begin{table}[t]
    \centering
    \setlength{\tabcolsep}{2.8pt}
    \caption{Per-class IoU comparison among Dedicated, Pretrained, and Adapted model on MCubeS Dataset.}
    \label{tab:mcubes-failure-case}
    \centering
    \begin{tabular}{lccccccccc}
        \toprule
        \multicolumn{1}{c}{} &
          \multicolumn{3}{c}{RGB} &
          \multicolumn{3}{c}{RGB+AoLP} &
          \multicolumn{3}{c}{RGB+AoLP+DoLP} \\
        \multicolumn{1}{c}{\multirow{-2}{*}{Class}} &
          Dedicated &
          Pretrained &
          Adapted &
          Dedicated &
          Pretrained &
          Adapted &
          Dedicated &
          Pretrained &
          Adapted \\
        \midrule
        \midrule
        Asphalt &
          85.68 &
          \textbf{69.74} &
          \textbf{85.80} &
          87.45 &
          83.20 &
          84.43 &
          87.02 &
          84.45 &
          84.14 \\
        Concrete &
          43.41 &
          39.18 &
          \textbf{43.72} &
          45.29 &
          43.28 &
          \textbf{45.36} &
          43.82 &
          41.20 &
          \textbf{45.57} \\
        Metal &
          51.36 &
          47.57 &
          \textbf{52.64} &
          53.11 &
          50.72 &
          \textbf{53.75} &
          50.65 &
          46.68 &
          \textbf{54.08} \\
        Road\_Marking &
          64.95 &
          67.33 &
          \textbf{73.79} &
          59.53 &
          72.63 &
          \textbf{74.50} &
          71.29 &
          72.76 &
          \textbf{74.58} \\
        Fabric &
          30.08 &
          26.88 &
          27.86 &
          30.52 &
          26.37 &
          30.42 &
          29.86 &
          25.20 &
          \textbf{30.50} \\
        Glass &
          51.20 &
          {\color[HTML]{FF0000} \textbf{44.56}} &
          50.95 &
          53.25 &
          51.91 &
          53.22 &
          50.88 &
          51.59 &
          \textbf{54.23} \\
        Plaster &
          0.11 &
          0.16 &
          \textbf{0.75} &
          0.37 &
          0.21 &
          \textbf{0.63} &
          0.37 &
          0.33 &
          \textbf{0.63} \\
        Plastic &
          21.81 &
          20.86 &
          \textbf{24.18} &
          23.22 &
          28.05 &
          \textbf{26.90} &
          22.05 &
          26.05 &
          \textbf{28.10} \\
        Rubber &
          26.11 &
          \textbf{15.24} &
          \textbf{30.40} &
          27.96 &
          {\color[HTML]{FF0000} \textbf{22.23}} &
          \textbf{28.76} &
          27.63 &
          28.76 &
          \textbf{30.15} \\
        Sand &
          59.65 &
          61.76 &
          \textbf{67.80} &
          58.75 &
          67.67 &
          \textbf{68.97} &
          62.88 &
          66.72 &
          \textbf{69.01} \\
        Gravel &
          64.24 &
          \textbf{36.73} &
          \textbf{72.86} &
          66.24 &
          63.40 &
          \textbf{69.25} &
          71.08 &
          {\color[HTML]{FF0000} \textbf{65.63}} &
          67.63 \\
        Ceramic &
          27.11 &
          {\color[HTML]{FF0000} \textbf{19.12}} &
          27.10 &
          30.11 &
          {\color[HTML]{FF0000} \textbf{22.70}} &
          24.83 &
          30.30 &
          25.97 &
          25.86 \\
        Cobblestone &
          68.76 &
          67.15 &
          68.06 &
          71.14 &
          67.48 &
          67.51 &
          71.25 &
          {\color[HTML]{FF0000} \textbf{66.47}} &
          67.80 \\
        Brick &
          41.34 &
          {\color[HTML]{FF0000} \textbf{35.96}} &
          \textbf{42.93} &
          41.90 &
          40.55 &
          \textbf{43.24} &
          43.52 &
          40.38 &
          \textbf{43.79} \\
        Grass &
          58.93 &
          {\color[HTML]{FF0000} \textbf{49.46}} &
          57.57 &
          56.89 &
          54.44 &
          \textbf{58.36} &
          56.77 &
          {\color[HTML]{FF0000} \textbf{49.92}} &
          \textbf{58.01} \\
        Wood &
          45.12 &
          \textbf{36.13} &
          \textbf{49.02} &
          44.99 &
          44.91 &
          \textbf{48.21} &
          44.64 &
          45.08 &
          \textbf{49.07} \\
        Leaf &
          76.62 &
          {\color[HTML]{FF0000} \textbf{71.55}} &
          74.91 &
          75.70 &
          73.90 &
          75.05 &
          75.60 &
          72.67 &
          75.03 \\
        Water &
          45.12 &
          {\color[HTML]{FF0000} \textbf{36.13}} &
          43.41 &
          41.46 &
          44.77 &
          \textbf{48.13} &
          52.39 &
          49.43 &
          \textbf{53.73} \\
        Human &
          4.27 &
          14.70 &
          \textbf{18.27} &
          3.32 &
          21.79 &
          \textbf{14.43} &
          1.33 &
          25.59 &
          \textbf{13.71} \\
        Sky &
          96.39 &
          {\color[HTML]{FF0000} \textbf{86.25}} &
          \textbf{96.47} &
          96.53 &
          96.04 &
          96.42 &
          96.34 &
          96.33 &
          \textbf{96.49} \\
        \midrule
        \textbf{Average} &
          48.11 &
          42.32 &
          \textbf{50.42} &
          48.39 &
          48.81 &
          \textbf{50.62} &
          49.48 &
          49.06 &
          \textbf{51.11} \\
        \bottomrule
    \end{tabular}
\end{table}

\begin{figure*}[t]
    \centering
    \subfloat[Visualization of predictions on MFNet dataset for multimodal semantic segmentation\label{fig:vis-mfnet-2}]{\includegraphics[width=0.95\textwidth]{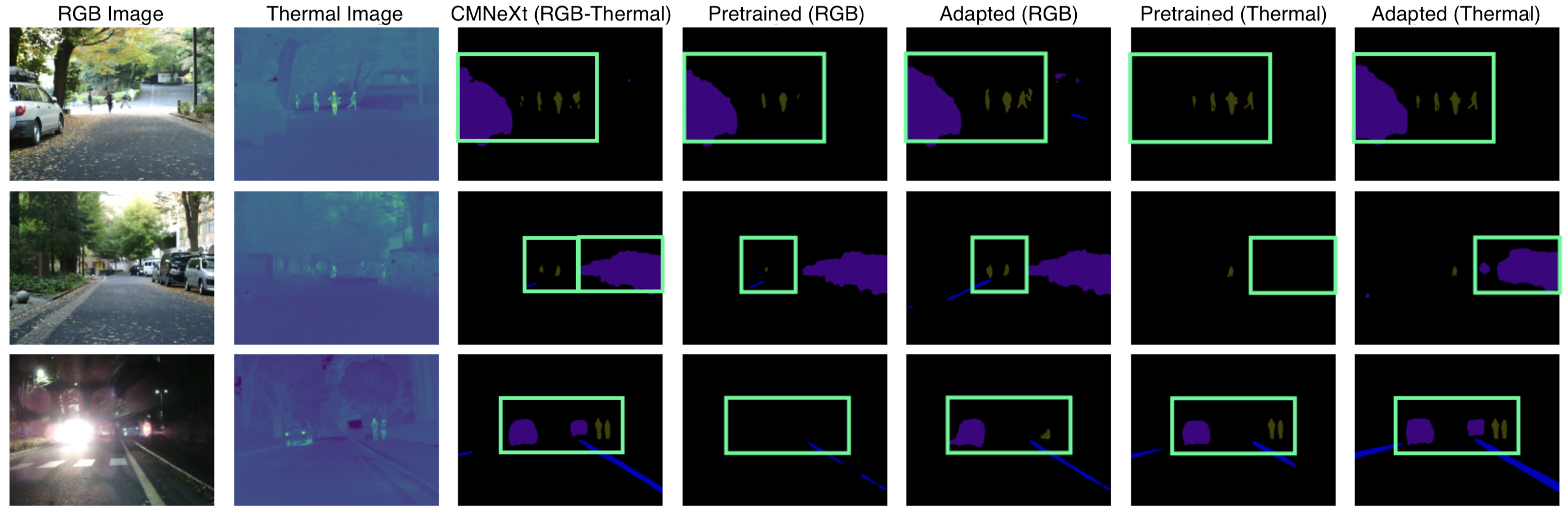}}
    
    \hfill
    \\[3.5pt]
    
    \subfloat[Visualization of predictions on NYUDv2 dataset for multimodal semantic segmentation\label{fig:vis-nyu-2}]{\includegraphics[width=0.95\textwidth]{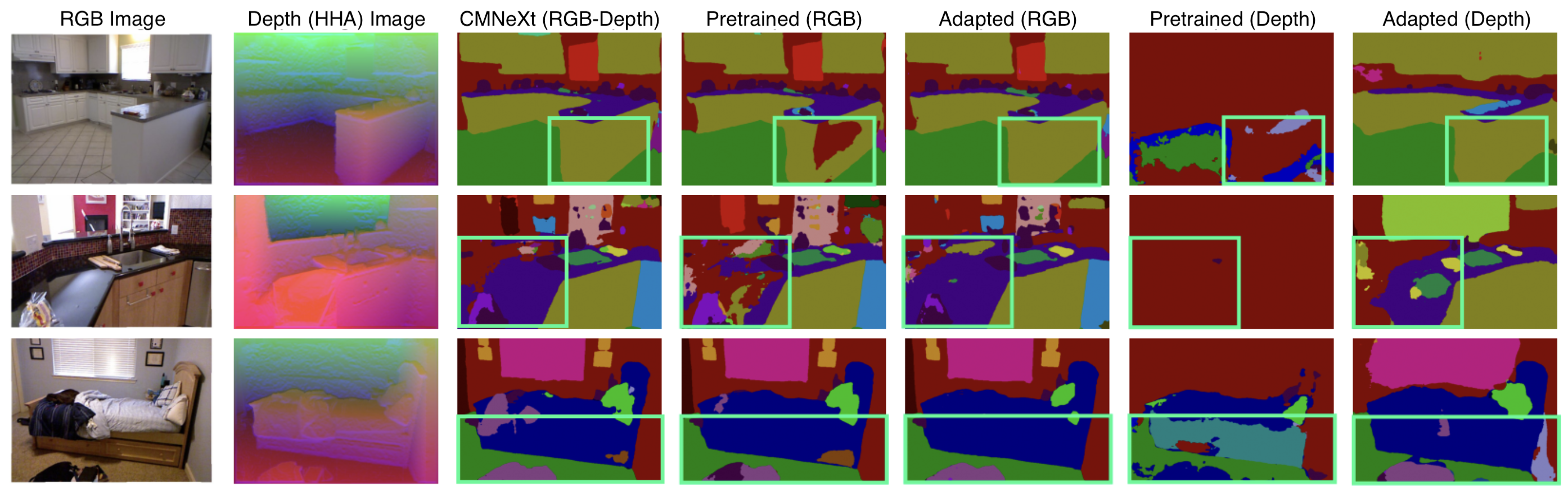}}
    
    \hfill
    \\[3.5pt]
    
    \subfloat[Visualization of predictions on MCubeS dataset for multimodal material segmentation\label{fig:vis-mcubes-2}]{\includegraphics[width=0.95\textwidth]{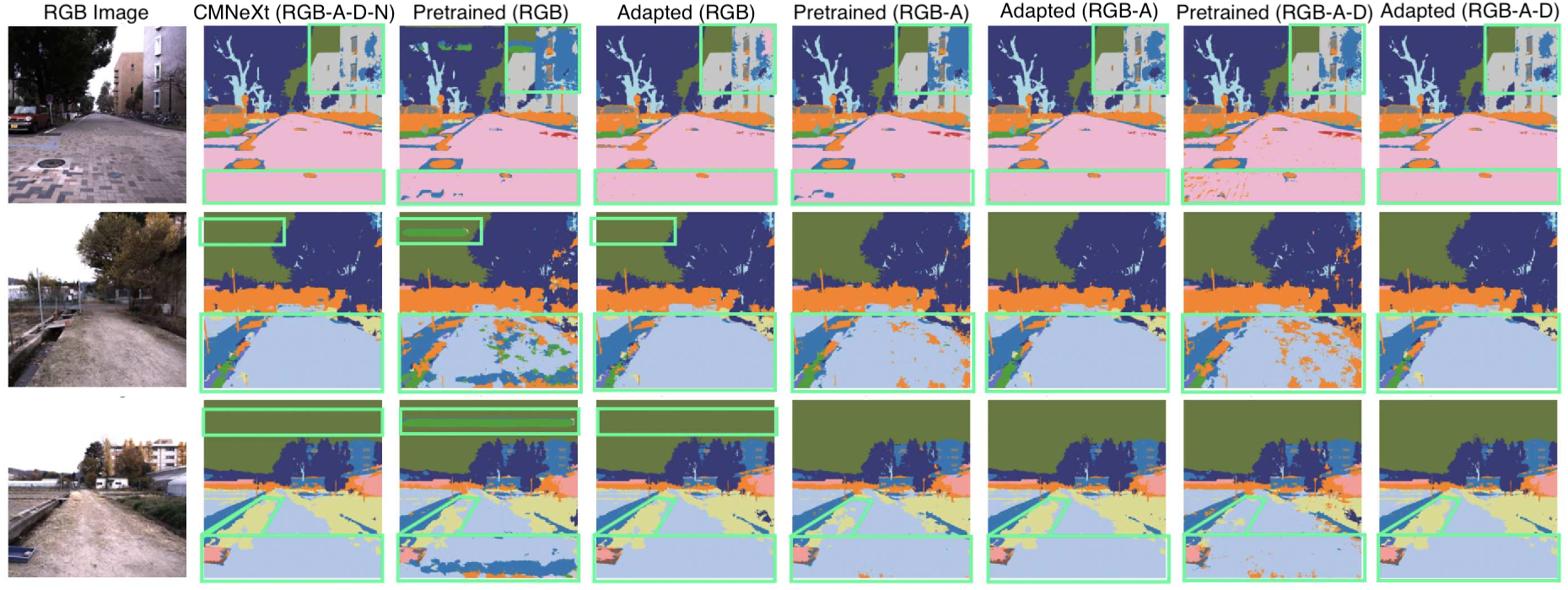}}

    \caption{Visualization of predicted segmentation maps for pretrained and adapted models on MFNet and NYUDv2 datasets for multimodal semantic segmentation and MCubeS dataset for multimodal material segmentation. Only RGB input images are shown from MCubeS dataset for brevity. CMNeXt column shows the predictions when all the modalities are available. Segmentation quality improves significantly after model adaptation for all the input modality combinations. A, D and N stand for angle of linear polarization, degree of linear polarization and near-infrared respectively.}
    \label{fig:vis-prediction-more}
\end{figure*}

Results for MCubeS dataset is shown on Table~\ref{tab:per-class-iou-mcubes-cmnext-vs-ours}. Here A, D, and N stand for angle of linear polarization (AoLP), degree of linear polarization (DoLP) and near-infrared (NIR) respectively. Experiments show that when only RGB is available and the rest of the modalities are missing, the adapted model performs better in detecting all the 20 classes present in the dataset. Gravel (36.2\%), asphalt (16.1\%), rubber (15.2\%), wood (12.9\%) and sky (10.2\%) are some of the classes who show the most performance boost after adaptation. In other input combinations, most of the classes see performance improvement compared to the pretrained model. Though we see some performance drop in a few classes, most of the classes show improvement in \% IoU which leads to the overall performance improvement after adaption.

\section{Cosine Similarity Analysis}
\label{sec:cos-sim-analysis-supp}
We show the cosine similarity for different missing modality scenarios on MCubeS dataset using CMNeXt as the base model in Figure~\ref{fig:mcubes-cos-sim}. These results show similar trends as discussed in Section~\ref{sec:mcubes-cosine-sim}, demonstrating a consistent increase in cosine similarity for all of the classes. This leads to an overall increase in performance for the adapted model compared to the pretrained model under various missing modality scenarios.

For multimodal action recognition, we utilize the UMDR \cite{zhou2023umdr} as the base pretrained model. When RGB is available but depth is missing, the adapted model demonstrates a significant increase in cosine similarity compared to the pretrained model, as illustrated in Figure~\ref{fig:ntu-cos-sim}. This enhancement translates to a 1.06\% improvement in overall performance, as shown in Table~\ref{tab:ntu-action-recognition}. When depth is available and RGB is missing, both the pretrained and adapted models show comparable cosine similarity. This is because the base UMDR model can handle depth-only data quite well and maintain a higher performance, resulting in similar performance metrics for both the pretrained and adapted models. 

This consistency across datasets and tasks strengthens the generalizability and effectiveness of the adaptation process in promoting model robustness to missing modalities.

\section{Effectiveness of our adaptation method on different backbones}
\label{sec:cmnext-with-different-backbones}
To assess the effectiveness of our method across different backbones, we replaced the MiT-B4 backbone in the CMNeXt model with ResNet-34, ResNet-50, and Swin-S. We use the same experimental setup as the original CMNeXt model \cite{zhang2023cmnext} while training the base models. In order to use the default window size and other configurations of Swin-S backbone, we resize the images to $448 \times 448$ for training and testing. For ResNets, we use the default image resolution of $640 \times 480$. Other hyperparameters are selected in the same manner as the original CMNeXt model. Then we adapt the pretrained base models for different missing modality scenarios using our approach. For adaptation, we use the same hyper-parameters as described in Section \ref{sec:impl-details} and \ref{sec:impl-details-supplementary}. 

We summarize the test set performance of the pre-trained and adapted CMNeXt models on MFNet dataset with different backbones in Table~\ref{tab:cmnext-different-backbones}. Our approach needs a small number of learnable parameters for each of the backbones. ResNet-34, ResNet-50, and Swin-S add only 0.37\%, 0.40\% and 0.33\% additional learnable parameters, respectively. Adapted CMNeXt with ResNet-34 shows 10.26 and 10.51 points improvement in mean accuracy (mAcc) and \%mIoU, respectively on average. Adapted CMNeXt with ResNet-50 provides an improvement of 18.54 and 15.36 points while the Adapted CMNeXt with Swin-S gains 18.49 and 16.63 points improvement in mAcc and \%mIoU, respectively on average. 

These results confirm that our adaptation approach generalizes across backbones, delivering significant performance gains with a small number of additional parameters.

\section{Why Dedicated Model Performs Better on MFNet Dataset?}
As shown in Table~\ref{tab:missing-modality-effect-on-cmnext}, the dedicated baseline performs better than adapted model on MFNet dataset. We further investigated the MFNet results and found that in the pretrained model a dominant modality (e.g., only RGB or Thermal input) contributes to the prediction of some specific classes as highlighted with gray background in Table~\ref{tab:mfnet-failure-case}. If the dominant modality is missing, a significant performance drop is observed for that class. For instance, for Person class Thermal modality dominates as RGB only provides 51.13\% IoU and Thermal only provides 68.47\% IoU; for Bike class RGB dominates as RGB only provides 61.94\% IoU whereas Thermal only provides 9.27\% IoU. The adapted model exhibits a similar pattern since it builds on the pretrained model. While adaptation improves the IoU for some classes significantly, it falls short in some cases. On the other hand, dedicated networks are trained independently with RGB or Thermal and are forced to make correct predictions with the respective modality only. That can be one possible reason why dedicated networks perform better than adapted for MFNet dataset. 

For other datasets like MCubeS , as shown in Table~\ref{tab:mcubes-failure-case}, the adapted model can improve performance on most of the classes and outperforms the dedicated network. We also observe an overall performance boost compared to the pretrained and dedicated networks.

The main drawback of the dedicated models is that we have to train as many independent models as the number of modality combinations. In contrast, our proposed method adapts a single pretrained model for different missing modality scenarios using a small number of additional parameters. Thus, our approach offers compute and memory efficiency while performing on par or better than dedicated models. 

\section{Visualization of Predicted Segmentation Maps}
We show the predicted segmentation maps from the pretrained and adapted models in Figure~\ref{fig:vis-prediction-more}. For each dataset, we show the input images, predictions from the base CMNeXt model when all the modalities are available, predictions from the adapted and pretrained models for different missing modality scenarios. For brevity, we only show RGB input images for MCubeS dataset. A, D and N stand for angle of linear polarization (AoLP), degree of linear polarization (DoLP) and near-infrared (NIR) respectively. Modalities that are available during testing are shown in parenthesis while other modalities are missing. 

For MFNet dataset, Figure~\ref{fig:vis-mfnet-2} shows that when only RGB is available, the pretrained model performs very poorly in detecting humans. On the other hand, if only thermal is available, the pretrained model can not detect cars very accurately. But the adapted model can detect both humans and cars more accurately in both of the scenarios. In all the cases, the predictions form the adapted model is closer to the predictions of the base CMNeXt model when all the modalities are available.

Predictions from NYUDv2 dataset is shown on Figure~\ref{fig:vis-nyu-2}. We can see that the adapted model can identify bed, furniture and other classes more accurately than the pretrained model for different missing modality scenarios. The pretrained model performs very poorly when only depth is available and RGB is missing. But detection accuracy improves significantly after model adaptation. For MCubeS dataset, as seen in Figure~\ref{fig:vis-mcubes-2}, predictions from the pretrained model shows artifacts when detecting different materials. On the other hand, the adapted model is showing more accuracy in detecting sky, cobblestone, sand and brick. For all the three datasets, the predictions from the adapted model is more accurate and closer to the all modality predictions of the base CMNeXt model.

\end{document}